\documentclass[11pt]{article} 
\usepackage{amssymb,amsfonts,amsmath,amscd}
\usepackage{latexsym} \usepackage{amscd} \usepackage{hhline}
\usepackage{graphicx} 
\usepackage[all]{xy} 
\usepackage{psfrag} 
\usepackage{pstricks,pst-node}
\usepackage{bbm}

\newtheorem{theorem}{Theorem}
\newtheorem{lemma}{Lemma}

\newlength{\minipagewidth}
\newcommand{\bookbox}[1]{
\par\bigskip\noindent
\begin{center}
\framebox[.9\textwidth]{
\begin{minipage}{0.9\minipagewidth}
{#1}
\end{minipage} }
\end{center}
\par\bigskip\noindent }

\def\e{e}

\def\P{\normalfont \texttt{Part}}

\def\ui{\underline{\boldsymbol{i}}}

\def\bt{{\mathbf{t}}}
\def\bj{\boldsymbol{j}}

\newcommand{\R}{\mathbb{R}}

\newcommand{\E}{E}  
\newcommand{\EXP}{\mathbb{E}} 

\def\d{\delta}
\def\b{\beta}
\def\eps{\varepsilon}
\def\wh{\widehat}
\def\ol{\overline}
\def\ti{\tilde}
\def\mc{\mathcal}

\def\cR{{\cal P}}
\def\bW{\ol{W}}

\newcommand{\PROB}{\mathbb{P}}

\newcommand{\IND}[1]{\mathbbm{1}_{\{ #1 \}}}
\newcommand{\I}{\mathbbm{1}}

\newcommand{\loss}{\ell}

\newcommand{\defeq}{\stackrel{\rm def}{=}}

\newcommand{\proof}{\medskip \par \noindent{\bf Proof.}\ }
\newcommand{\proofof}[1]{\medskip \par \noindent{\bf Proof of #1.}\ }

\newcommand{\blackslug}{$\Box$}
\newcommand{\qed}{\quad\blackslug\lower 8.5pt\null\par\medskip}
\newcommand{\QED}{\text{\quad\blackslug\lower 8.5pt\null}}

\newcommand{\C}{\mathcal{C}}

\newcommand{\bB}{\boldsymbol{B}}

\newcommand{\bi}{\boldsymbol{i}}
\newcommand{\bI}{\boldsymbol{I}}

\newcommand{\bp}{\boldsymbol{p}}

\newcommand{\bv}{\ol{v}}
\newcommand{\bw}{\ol{w}}

\newcommand{\bb}{\boldsymbol{b}}

\newcommand{\bl}{\boldsymbol{\ell}}

\newcommand{\figscale}[2]{\includegraphics[scale=#2,clip=false]{#1}}

\newcommand\wt{\widetilde}
\newcommand{\tL}{\wt{L}}
\newcommand{\tl}{\ti{\loss}}

\def\cP{{\cal P}}
\def\eps{\epsilon}

\allowdisplaybreaks
\begin{document}
\bibliographystyle{plain}
\title{The on-line shortest path problem under partial monitoring}

\author{Andr\'as Gy\"orgy \qquad Tam\'as Linder \qquad G\'abor
Lugosi \\ Gy\"orgy Ottucs\'ak
\thanks{A. Gy\"orgy is with the Machine Learning Research Group,
Computer and Automation Research Institute of the Hungarian Academy
of Sciences, Kende u.\ 11-13, Budapest, Hungary, H-1111 (email:
\texttt{gya@szit.bme.hu}). T. Linder is with the Department of
Mathematics and Statistics, Queen's University, Kingston, Ontario,
Canada K7L 3N6 (email: \texttt{linder@mast.queensu.ca}). G. Lugosi
is with ICREA and Department of Economics, Universitat Pompeu Fabra,
Ramon Trias Fargas 25-27, 08005 Barcelona, Spain (email:
\texttt{gabor.lugosi@gmail.com}). Gy. Ottucs\'ak is with Department
of Computer Science and Information Theory, Budapest University of
Technology and Economics, Magyar Tud\'osok K\"or\'utja 2., Budapest,
Hungary, H-1117. He is also with the Machine Learning Research Group,
Computer and Automation Research Institute of the Hungarian Academy of Sciences
(email: \texttt{oti@szit.bme.hu}).
\newline
 This research was
supported in part by the J\'anos Bolyai Research Scholarship of the
Hungarian Academy of Sciences, the Mobile Innovation Center of
Hungary, by the Natural Sciences and Engineering Research Council
(NSERC) of Canada, by the Spanish Ministry of Science and Technology
grant MTM2006-05650, by the PASCAL Network of Excellence
under EC grant no.\ 506778 and by the High Speed Networks Laboratory,
Department of Telecommunications and Media Informatics,
Budapest University of Technology and Economics.
Parts of this paper have been presented at COLT'06.
}}
\maketitle

\thispagestyle{empty}
\begin{abstract}
The on-line shortest path problem is considered under various models
of partial monitoring.  Given a weighted directed acyclic graph whose
edge weights can change in an arbitrary (adversarial) way, a decision
maker has to choose in each round of a game a path between two
distinguished vertices such that the loss of the chosen path (defined
as the sum of the weights of its composing edges) be as small as
possible.  In a setting generalizing the multi-armed bandit problem,
after choosing a path, the decision maker learns only the weights of
those edges that belong to the chosen path. For this problem, an
algorithm is given whose average cumulative loss in $n$ rounds exceeds
that of the best path, matched off-line to the entire sequence of the
edge weights, by a quantity that is proportional to $1/\sqrt{n}$ and
depends only polynomially on the number of edges of the graph. The
algorithm can be implemented with linear complexity in the number of
rounds $n$ and in the number of edges.  An extension to the so-called
label efficient setting is also given, in which the decision maker is
informed about the weights of the edges corresponding to the chosen
path at a total of $m \ll n$ time instances.  Another extension is
shown where the decision maker competes against a time-varying path, a
generalization of the problem of tracking the best expert. A version
of the multi-armed bandit setting for shortest path is also discussed
where the decision maker learns only the total weight of the chosen path
but not the weights of the individual edges on the path. Applications to
routing in packet switched networks along with simulation results are
also presented.
\end{abstract}

\medskip
\noindent
{\bf Index Terms:} On-line learning, shortest path problem, multi-armed bandit problem.
\thispagestyle{empty}
\newpage

\section{Introduction}
In a sequential decision problem, a decision maker (or forecaster)
performs a sequence of actions. After each action the decision maker
suffers some loss, depending on the response (or state) of the
environment, and its goal is to minimize its cumulative loss over a
certain period of time. In the setting considered here, no
probabilistic assumption is made on how the losses corresponding to
different actions are generated. In particular, the losses may depend
on the previous actions of the decision maker, whose goal is to
perform well relative to a set of reference forecasters (the so-called
``experts'') for any possible behavior of the environment. More
precisely, the aim of the decision maker is to achieve asymptotically
the same average (per round) loss as the best expert.

Research into this problem started in the 1950s (see, for example, 
Blackwell \cite{Bla56} and Hannan \cite{Han57} for some of the basic
results) and gained new life in the 1990s following the work of Vovk
\cite{Vov90}, Littlestone and Warmuth \cite{LiWa94}, and
Cesa-Bianchi \emph{et al$.$} \cite{CeFrHaHeScWa97}. These results show
that for any bounded loss function, if the decision maker has access
to the past losses of all experts, then it is possible to construct
on-line algorithms that perform, for any possible behavior of the
environment, almost as well as the best of $N$ experts. More precisely,
the per round cumulative loss of these algorithms
is at most as large as that of the best expert plus a quantity
proportional to $\sqrt{\ln N/n}$ for any bounded loss function,
where $n$ is the number of rounds in the decision game. The
logarithmic dependence on the number of experts makes it possible to
obtain meaningful bounds even if the pool of experts is very large.

In certain situations the decision maker has only limited knowledge
about the losses of all possible actions. For example, it is often
natural to assume that the decision maker gets to know only the loss
corresponding to the action it has made, and has no information
about the loss it would have suffered had it made a different
decision. This setup is referred to as the \emph{multi-armed bandit
problem}, and was considered, in the adversarial setting,
 by Auer \emph{et al$.$}  \cite{AuCeFrSc02}
who gave an algorithm
whose normalized regret (the difference of the algorithm's average
loss  and that of the best expert) is upper bounded by a quantity 
which is  proportional to $\sqrt{N\ln N/n}$. Note that, compared to the
\emph{full information}  case  described above where the losses of all
possible actions are revealed to the decision maker, there is an
extra $\sqrt{N}$ factor  in the performance bound, which seriously
limits the usefulness of the bound if the number of experts is
large.

Another interesting example for the limited information case is the
so-called \emph{label efficient decision problem} (see Helmbold and
Panizza \cite{HP97}) in which it is
too costly to observe the state of the environment, and so the
decision maker can query the losses of all possible actions for only
a limited number of times. A recent result of Cesa-Bianchi, Lugosi,
and Stoltz \cite{CeLuSt05} shows that in this case, if the decision
maker can query the losses $m$ times during a period of length $n$,
then it can achieve $O(\sqrt{\ln N/m})$ average excess loss relative
to the best expert.

In many applications the set of experts has a certain structure that
may be exploited to construct efficient on-line decision algorithms.
The construction of such algorithms has been of great interest in
computational learning theory. A partial list of works dealing with
this problem includes Herbster and Warmuth~\cite{HeWa98},
Vovk~\cite{Vov99}, Bousquet and Warmuth~\cite{BoWa02}, Helmbold and
Schapire~\cite{HeSc97}, Takimoto and Warmuth~\cite{TaWa03}, 
Kalai and Vempala~\cite{KaVe03}, Gy\"orgy \emph{at
al$.$}~\cite{GyLiLu04,GyLiLu04a,GyLiLu05a}. For a more complete
survey, we refer to Cesa-Bianchi and Lugosi \cite[Chapter 5]{CeLu06}.

In this paper we study the on-line shortest path problem, a
representative example of structured expert classes that has
received attention in the literature for its many applications,
including, among others, routing in communication networks; see,
e.g., Takimoto and Warmuth \cite{TaWa03},  Awerbuch \emph{et al}.\
\cite{AwHoRuKl05}, or Gy\"orgy and Ottucs\'ak \cite{GyOt06}, and
adaptive quantizer design in zero-delay lossy source coding; see,
Gy\"orgy \emph{et al$.$}~\cite{GyLiLu04,GyLiLu04a,GyLiLu05d}.
In this problem, a weighted directed (acyclic) graph is given whose
edge weights can change in an arbitrary manner, and the decision
maker has to pick in each round a path between two given vertices,
such that the weight of this path (the sum of the weights of its
composing edges) be as small as possible.

Efficient solutions, with time and space complexity proportional to
the number of edges rather than to the number of paths (the latter
typically being exponential in the number of edges), have been given
in the full information case, where in each round the weights of all
the edges are revealed after a path has been chosen; see, for example,
Mohri \cite{moh98}, Takimoto and Warmuth \cite{TaWa03}, Kalai and
Vempala \cite{KaVe03}, and Gy\"orgy \emph{et al$.$}~\cite{GyLiLu05a}.

In the bandit setting only the weights of the edges or just
the sum of the weights of the edges composing the chosen path are
revealed to the decision maker. If one applies the general bandit
algorithm of Auer \emph{et al}.\ \cite{AuCeFrSc02}, the
resulting bound will be too large to be of practical use because of
its square-root-type dependence on the number  of paths $N$. On the
other hand, using the special graph structure in the problem,
Awerbuch and Kleinberg \cite{AwKl04} and McMahan and Blum
\cite{McBl04}
managed to get rid of the exponential dependence on the number of
edges in the performance bound. They achieved this by extending the
exponentially weighted average predictor and the 
follow-the-perturbed-leader algorithm of Hannan \cite{Han57} to the
generalization of the multi-armed bandit setting for shortest paths,
when only the sum of the weights of the edges is available for the
algorithm.  However, the dependence of the bounds obtained in
\cite{AwKl04} and \cite{McBl04} on the number of rounds $n$ is
significantly worse than the $O(1/\sqrt{n})$ bound of Auer \emph{et
al}.\ \cite{AuCeFrSc02}.  Awerbuch and Kleinberg \cite{AwKl04}
consider the model of ``non-oblivious'' adversaries for shortest path
(i.e., the losses assigned to the edges can depend on the previous
actions of the forecaster) and prove an $O(n^{-1/3})$ bound for the
expected per-round regret. McMahan and Blum \cite{McBl04} give a
simpler algorithm than in \cite{AwKl04} however obtain a bound of the
order of $O(n^{-1/4})$ for the expected regret.

In this paper we provide an extension of the bandit algorithm of
Auer \emph{et al}.\  \cite{AuCeFrSc02} unifying the advantages of the
above approaches, with a performance bound that is polynomial in
the number of edges, and converges to zero at the right
$O(1/\sqrt{n})$ rate as the number of rounds increases. We
achieve this bound in a model which assumes that the losses of all
edges on the path chosen by the forecaster are available separately
after making the decision. We also discuss the case (considered by
\cite{AwKl04} and \cite{McBl04}) in which only the total loss (i.e.,
the sum of the losses on the chosen path) is known to the decision
maker. We exhibit a simple algorithm which achieves an $O(n^{-1/3})$
per-round regret \emph{with high probability} against ``non-oblivious''
adversary.
In this case it remains an open problem to find an algorithm
whose cumulative loss is polynomial in the number of edges of the graph
and decreases as $O(n^{-1/2})$ with the number of rounds.

In Section~\ref{sec:path} we formally  define the on-line shortest path
 problem, which is extended to the multi-armed bandit
 setting in Section~\ref{sec:bandit}. Our new algorithm for the shortest path
 problem in the bandit setting is given in Section~\ref{sec:alg}
together with its performance
 analysis. The algorithm is extended to solve the shortest path problem in a
 combined label efficient multi-armed bandit setting in
 Section~\ref{sec:leband}. Another extension, when the algorithm competes
 against a time-varying path is studied in Section \ref{sec:tracking}.
 An algorithm for the ``restricted'' multi-armed bandit setting
(when only the sums of the losses of the edges are available) is given
 in Section~\ref{sec:generalizedbandit}.
Simulation results are presented in Section~\ref{sec:sim}.

\section{The shortest path problem}
\label{sec:path}

Consider a network represented by a set of vertices connected by
edges, and assume that we have to send a stream of packets from a
distinguished vertex, called {\em source}, to another distinguished vertex,
called {\em destination}. At each time slot a packet is sent along a chosen
route connecting source and destination.  Depending on the traffic,
each edge in the network may have a different delay, and the total
delay the packet suffers on the chosen route is the sum of delays of
the edges composing the route. The delays may change from one time
slot to the next one in an arbitrary way, and our goal is to find a
way of choosing the route in each time slot such that the sum of the
total delays over time is not significantly more than that of the best
fixed route in the network. This adversarial version of the routing
problem is most useful when the delays on the edges can change
dynamically, even depending on our previous routing decisions. This is
the situation in the case of ad-hoc networks, where the network
topology can change rapidly, or in certain secure networks, where the
algorithm has to be prepared to handle denial of service attacks, that
is, situations where willingly malfunctioning vertices and links
increase the delay; see, e.g., Awerbuch \emph{et al}.\
\cite{AwHoRuKl05}.

This problem can  be cast  naturally as a sequential decision problem
in which each possible route is represented by an action. However,
the number of routes is typically exponentially large in the number
of edges, and therefore computationally efficient algorithms are
called for. Two solutions of different flavor have been
proposed. One of them is based on a follow-the-perturbed-leader
forecaster, see Kalai and Vempala \cite{KaVe03},
 while the other is
based on an efficient computation of the exponentially weighted
average forecaster, see, for example, Takimoto and Warmuth
\cite{TaWa03}. Both solutions have different advantages and may be
generalized in different directions.

\begin{figure}[ht]
\begin{center}
\leavevmode
\psfrag{u}{$u$}
\psfrag{v}{$v$}
\figscale{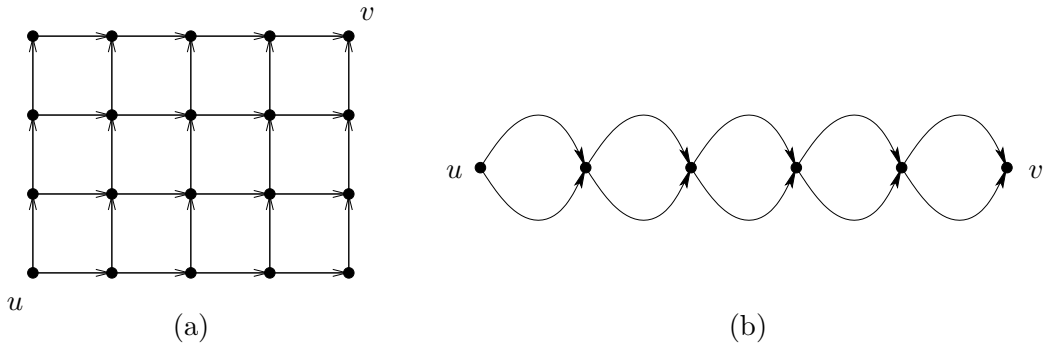}{0.7}
\end{center}
\caption{
\label{f:dags}
Two examples of directed acyclic graphs for the shortest path problem.
}
\vspace{-1.4cm}\hspace{3.6cm}(a)\hspace{6.9cm}(b)\vspace{1.4cm}
\end{figure}

To formalize the problem, consider a (finite) directed acyclic graph
with a set of edges $E=\{e_1,\ldots,e_{|E|}\}$ and a set of vertices
$V$. Thus, each edge $e \in E$ is an ordered pair of vertices
$(v_1,v_2)$.  Let $u$ and $v$ be two distinguished vertices in $V$.
A {\em path} from $u$ to $v$ is a sequence of edges
$e^{(1)},\ldots,e^{(k)}$ such that $e^{(1)}=(u,v_1)$,
$e^{(j)}=(v_{j-1},v_j)$ for all $j=2,\ldots,k-1$, and
$e^{(k)}=(v_{k-1},v)$.
Let $\cP=\{\bi_1,\ldots,\bi_N\}$ denote the set of all such
paths. For simplicity, we assume that every edge in $E$ is on some
path from $u$ to $v$ and every vertex in $V$ is an endpoint of an
edge (see Figure~\ref{f:dags} for examples).

In each round $t=1,\ldots,n$ of the decision game, the decision
maker chooses a path $\bI_t$ among all paths from $u$ to $v$. Then a
loss $\ell_{e,t} \in [0,1]$ is assigned to each edge $e \in E$. We
write $e \in \bi$ if the edge $e\in E$ belongs to the path $\bi \in
\cP$, and with a slight abuse of notation the loss of a path $\bi$
at time slot $t$ is also represented by $\ell_{\bi,t}$.
Then $\ell_{\bi,t}$ is given as
\[
   \ell_{\bi,t} = \sum_{e \in \bi} \ell_{e,t}
\]
and therefore the cumulative loss up to time $t$ of each path $\bi$ takes the
additive form
\[
   L_{\bi,t}
       = \sum_{s=1}^t \ell_{\bi,s}
       = \sum_{e \in \bi} \sum_{s=1}^t \ell_{e,s}
\]
where the inner sum on the right-hand side is the loss accumulated
by edge $e$ during the first $t$ rounds of the game. The cumulative
loss of the algorithm is
\[
\wh{L}_t
       = \sum_{s=1}^t \ell_{\bI_s,s}
       = \sum_{s=1}^t \sum_{e \in \bI_s} \ell_{e,s}~.
\]

It is well known that for a general loss sequence, the decision
maker must be allowed to use randomization to be able to approximate the
performance of the best expert (see, e.g., Cesa-Bianchi and Lugosi
\cite{CeLu06}). Therefore, the path $\bI_t$ is chosen randomly according to
some distribution $\bp_t$ over all paths from $u$ to $v$. We study
the normalized regret over $n$ rounds of the game 
\[
  \frac{1}{n}\left(\wh{L}_n - \min_{\bi \in \cP} L_{\bi,n}\right)
\]
where the minimum is taken over all paths $\bi$ from $u$ to $v$.

For example, the exponentially weighted average forecaster
(\cite{Vov90}, \cite{LiWa94}, \cite{CeFrHaHeScWa97}), calculated over
all possible paths, has regret
\[
\frac{1}{n}\left(\wh{L}_n - \min_{\bi \in \cP} L_{\bi,n} \right)
 \le K\left(\sqrt{\frac{\ln N}{2n}}
      + \sqrt{\frac{\ln (1/\delta)}{2n}} \right)
\]
with probability at least $1-\delta$, where $N$ is the total number
of paths from $u$ to $v$ in the graph and $K$ is the length of the
longest path.

\section{The multi-armed bandit setting}
\label{sec:bandit}

In this section we discuss the ``bandit'' version of the shortest path
problem. In this setup, which is more realistic in many applications,
the decision maker has only access to the losses corresponding to the
paths it has chosen. For example, in the routing problem this means
that information is available on the delay of the route the packet is
sent on, and not on other routes in the network.

We distinguish between two types of bandit problems,
both of which are natural generalizations of the simple bandit problem to the
shortest path problem.
In the first variant, the decision maker has access to the losses of
those edges that are on the path it has chosen. That is, after
choosing a path $\bI_t$ at time $t$, the value of the loss
$\ell_{e,t}$ is revealed to the decision maker if and only if $e \in
\bI_t$. We study this case and its extensions in Sections
\ref{sec:alg}, \ref{sec:leband}, and \ref{sec:tracking}.

The second variant is a more restricted version in which the
loss of the chosen path is observed, but no information is available on the
individual  losses of the edges belonging to the path. That is, after
choosing a path $\bI_t$ at time $t$, only the value of the loss of
the path $\ell_{\bI_t,t}$ is revealed to the decision maker. Further
on we call this setting as the \emph{restricted} bandit problem for
shortest path. We consider this restricted problem in Section
\ref{sec:generalizedbandit}.

Formally, the on-line shortest path problem in the multi-armed
bandit setting is described as follows: at each time instance
$t=1,\ldots,n$, the decision maker picks a path $\bI_t \in \cP$ from
$u$ to $v$. Then the environment assigns loss $\ell_{e,t} \in [0,1]$
to each edge $e \in E$, and the decision maker suffers loss
$\ell_{\bI_t,t}=\sum_{e \in \bI_t} \ell_{e,t}$. In the unrestricted case
the losses $\ell_{e,t}$ are revealed for all $e \in \bI_t$, while  in the
restricted case only $\ell_{\bI_t,t}$ is revealed. Note that in both
cases $\ell_{e,t}$ may depend on $\bI_1,\ldots,\bI_{t-1}$, the
earlier choices of the decision maker.

For the basic multi-armed bandit problem, Auer \emph{et al}.\
\cite{AuCeFrSc02} gave an algorithm, based on exponential weighting
with a biased estimate of the gains (defined, in our case, as
$g_{\bi,t}=K-\ell_{\bi,t}$), combined with uniform exploration.
Applying their algorithm to the on-line shortest path problem in the
bandit setting results in a performance that can be bounded,
for any $0<\delta<1$ and fixed time
horizon $n$, with
probability at least $1-\delta$, by
\[
  \frac{1}{n}\left(\wh{L}_n - \min_{\bi \in \cP} L_{\bi,n}\right)
\le    \frac{11K}{2}\sqrt{\frac{N\ln(N/\delta)}{n}} + \frac{K\ln
N}{2n}~.
\]
(The constants follow from a slightly improved version; see Cesa-Bianchi
and Lugosi \cite{CeLu06}.)

However, for the shortest path problem this bound is unacceptably
large because, unlike in the full information case, here the
dependence on the number of all paths $N$ is not merely
logarithmic, while $N$ is typically exponentially large in the size of the graph
(as in the two simple examples of Figure~\ref{f:dags}). 
In order to achieve a bound that does not grow
exponentially with the number of edges of the graph, it is imperative
to make use of the dependence structure of the losses of the different
actions (i.e., paths).  Awerbuch and Kleinberg~\cite{AwKl04} and
McMahan and Blum~\cite{McBl04} do this by extending low complexity
predictors, such as the follow-the-perturbed-leader forecaster
\cite{Han57}, \cite{KaVe03} to the restricted bandit setting.
However, in both cases the price to pay for the polynomial dependence
on the number of edges is a worse dependence on the length $n$ of the
game.

\section{A bandit algorithm for shortest paths}
\label{sec:alg}

In this section we describe a variant of the bandit algorithm of
\cite{AuCeFrSc02} which achieves the desired performance for the
shortest path problem. The new algorithm uses the fact that when the
losses of the edges of the chosen path are revealed, then this also
provides some information about the losses of each path sharing common
edges with the chosen path.

For each edge $e \in E$, and $t=1,2,\ldots$,
introduce the \emph{gain} $g_{e,t} =1-\ell_{e,t}$,
and for each path $\bi \in \cP$,
let the gain be the sum of the gains of the edges on the
path, that is,
\[
g_{\bi,t}=\sum_{e \in \bi} g_{e,t}~.
\]
The conversion from losses to gains is done in order to facilitate
the subsequent performance analysis.  To
simplify the conversion, we assume that each path $\bi \in \cP$ is
of the same length $K$ for some $K>0$.  Note that although this
assumption may seem to be restrictive at the first glance, from each
acyclic directed graph $(V,E)$ one can construct a new graph by
adding at most $(K-2)(|V|-2)+1$ vertices and edges (with constant
weight zero) to the graph without modifying the weights of the paths
such that each path from $u$ to $v$ will be of length $K$, where $K$
denotes the length of the longest path of the original graph. If
the number of edges is quadratic in the number of vertices,
the size of the graph is not increased substantially.

A main feature of the algorithm below is that the gains are
estimated for each edge and not for each path. This modification
results in an improved upper bound on the performance with the
number of edges in place of the number of paths. Moreover, using
dynamic programming as in Takimoto and Warmuth \cite{TaWa03}, the
algorithm can be computed efficiently. Another important ingredient
of the algorithm is that one needs to make sure that every edge is
sampled sufficiently often. To this end, we introduce a set $\C$ of
\emph{covering paths} with the property that for each edge $e \in E$
there is a path $\bi \in \C$ such that $e \in \bi$. Observe that one
can always find such a covering set of cardinality $|\C| \le |E|$.

We note that the algorithm of \cite{AuCeFrSc02} is a special case of
the algorithm below: For any multi-armed bandit problem with $N$
experts, one can define a graph with two vertices $u$ and $v$, and
$N$ directed edges from $u$ to $v$ with weights corresponding to the
losses of the experts. The solution of the shortest path problem in
this case is equivalent to that of the original bandit problem  with
choosing expert $\bi$ if the corresponding edge is chosen. For this
graph, our algorithm reduces to the original algorithm of
\cite{AuCeFrSc02}.

\bookbox{
\begin{center}
A BANDIT ALGORITHM FOR SHORTEST PATHS
\end{center}
\textbf{Parameters:} real numbers $\beta>0$, $0 < \eta,\gamma < 1$.
\\
\textbf{Initialization:} Set $w_{e,0}=1$ for each $e \in E$,
$\bw_{\bi,0}=1$ for each $\bi \in \cP$, and $\ol{W}_0=N$. For
each round $t=1,2,\ldots$
\vspace{-0.23cm}\begin{itemize}
\item[(a)] Choose a path $\bI_t$ at random according to the
  distribution $\bp_t$ on $\cP$, defined by
\[
p_{\bi,t}=\begin{cases} (1-\gamma)\frac{\bw_{\bi,t-1}}{\ol{W}_{t-1}}
+ \frac{\gamma}{|\C|}
& \text{ if $\bi \in \C$} \\
(1-\gamma)\frac{\bw_{\bi,t-1}}{\ol{W}_{t-1}} & \text{ if $\bi
\not\in \C$.}
\end{cases}
\]
\vspace{-0.52cm}
\item[(b)] Compute the probability of choosing each edge $e$ as
\begin{eqnarray*}
q_{e,t} = \sum_{\bi:e\in\bi} p_{\bi,t} =
(1-\gamma)\frac{\sum_{\bi:e\in\bi} \bw_{\bi,{t-1}}}{\ol{W}_{t-1}}+
\gamma\frac{\left|\{\bi \in \C:e\in\bi\}\right|}{|\C|}.
\end{eqnarray*}
\vspace{-0.72cm}
\item[(c)] Calculate the estimated gains \vspace{-0.12cm}
\[
g'_{e,t}=\begin{cases}
\frac{g_{e,t}+\beta}{q_{e,t}} & \text{ if $e \in \bI_t$} \\
\frac{\beta}{q_{e,t}} & \text{ otherwise.}
\end{cases}
\]
\vspace{-0.52cm}
\item[(d)] Compute the updated weights \vspace{-0.18cm}
\begin{eqnarray*}
w_{e,t}&=&w_{e,t-1} e^{\eta g'_{e,t}} \\
\bw_{\bi,t}&=&\prod_{e \in \bi} w_{e,t}= \bw_{\bi,t-1} e^{\eta
g'_{\bi,t}}
\end{eqnarray*} \vspace{-0.58cm}

\noindent where $g'_{\bi,t}=\sum_{e\in\bi}g'_{e,t}$, and the sum of the total
weights of the paths \vspace{-0.14cm}
\[
\ol{W}_t=\sum_{\bi \in \cP} \bw_{\bi,t}.
\]
\end{itemize}\vspace{-0.44cm}
}

The main result of the paper is the following performance bound for
the shortest-path bandit algorithm. It states that the per round
regret of the algorithm, after $n$ rounds of play,
is, roughly, of the order of $K\sqrt{|E|\ln N/n}$
where $|E|$ is the number of edges of the graph, $K$ is the length
of the paths, and $N$ is the total number of paths.

\begin{theorem}
\label{th:bandit} For any $\delta \in (0,1)$ and parameters
$0\le\gamma<1/2$, $0<\beta\le 1$, and $\eta>0$ satisfying $2\eta
K|\C|\le \gamma$, the performance of the algorithm defined above can
be bounded, with probability at least $1-\delta$, as
\[
\frac{1}{n}\left( \wh{L}_n-\min_{\bi\in\cP}
L_{\bi,n}\right)
\le K\gamma+2\eta K^2|\C|
+\frac{K}{n\beta}\ln\frac{|E|}{\delta}+\frac{\ln N}{n\eta} + |E
|\beta.
\]
In particular, choosing
$\beta=\sqrt{\frac{K}{n|E|}\ln\frac{|E|}{\delta}}$, $\gamma=2 \eta K
|\C|$, and $\eta=\sqrt{\frac{\ln N}{4nK^2|\C|}}$ yields for all
$n\ge \max\left\{\frac{K}{|E |}\ln\frac{|E|}{\delta}, 4|\C|\ln
N\right\}$,
\[
\frac{1}{n}\left( \wh{L}_n-\min_{\bi\in\cP}
L_{\bi,n}\right)
\le 
2\sqrt{\frac{K}{n}}\left(\sqrt{4K|\C|\ln N}+ \sqrt{|E |\ln\frac{|E
|}{\delta}}\; \right).
\]
\end{theorem}%

The proof of the theorem is based on the analysis of the original
algorithm of \cite{AuCeFrSc02} with necessary modifications required
to transform parts of the argument from paths to edges, and to
use the connection between the gains of paths sharing common
edges.

For the analysis we introduce some notation:
\[
G_{\bi,n}=\sum_{t=1}^n g_{\bi,t} \qquad \text{and} \qquad
G'_{\bi,n}=\sum_{t=1}^n g'_{\bi,t}
\]
for each $\bi \in \cR$ and
\[
G_{\e,n}=\sum_{t=1}^n g_{\e,t} \qquad \text{and} \qquad
G'_{\e,n}=\sum_{t=1}^n g'_{\e,t}
\]
for each $\e \in \E$, and
\[
\wh{G}_n = \sum_{t=1}^n  g_{\bI_t,t}.
\]

The following lemma, shows
that the deviation of the true cumulative gain from the estimated
cumulative gain is of the order of $\sqrt{n}$. The proof is a
modification of \cite[Lemma 6.7]{CeLu06}.

\begin{lemma}
\label{le:supermartingale2} For any $\delta \in (0,1)$, $0 \le \beta
< 1$ and $e \in E $ we have
\[ \PROB\left[G_{e,n}>G'_{e,n}+\frac{1}{\beta}\ln\frac{|E
|}{\delta}\right] \le \frac{\delta}{|E |}~.\] 
\end{lemma}
\proof Fix $e\in E $. For any $u>0$ and $c>0$, by the Chernoff bound we have
\begin{equation}
\label{eq:chernoff} \PROB[G_{\e,n}>G'_{\e,n}+u]\le e^{-cu} \EXP
e^{c(G_{\e,n}-G'_{\e,n})}~.
\end{equation}
Letting $u=\ln(|E |/\d)/\b$ and $c=\b$, we get
\begin{align*}
e^{-cu} \EXP e^{c(G_{\e,n}-G'_{\e,n})}
=e^{-\ln(|\E|/\d)}\EXP e^{\b(G_{\e,n}-G'_{\e,n})}
= \frac{\d}{|\E|} \EXP e^{\b(G_{\e,n}-G'_{\e,n})}~,
\end{align*}
so it suffices to prove that $\EXP e^{\beta(G_{e,n}-G'_{e,n})}\le 1$ for all $n$.
To this end, introduce
\[
Z_t=e^{\beta(G_{e,t}-G'_{e,t})}~.
\]
Below we show that $\EXP_t[Z_t] \le Z_{t-1}$ for $t \ge 2$ where
$\EXP_t$ denotes the conditional expectation
$\EXP[\cdot|\bI_1,\ldots,\bI_{t-1}]$~.
Clearly,
\[
Z_t=Z_{t-1} \exp\left(\beta\left(g_{e,t}-\frac{\IND{e\in
\bI_t}g_{e,t}+\beta}{q_{e,t}}\right)\right)~.
\]
Taking conditional expectations, we obtain
\begin{eqnarray}
\lefteqn{\EXP_t[ Z_t ]} \nonumber \\*
&=&Z_{t-1} \EXP_t\left[\exp\left(\beta\left(g_{e,t}-\frac{\IND{e\in
    \bI_t}g_{e,t}+\beta}{q_{e,t}}\right)\right)\right]
        \nonumber \\
&=&Z_{t-1} e^{-\frac{\beta^2}{q_{e,t}}}
        \EXP_t\left[\exp\left(\beta\left(g_{e,t}-\frac{\IND{e\in
    \bI_t}g_{e,t}}{q_{e,t}}\right)\right)\right]
        \nonumber \\
&\le&Z_{t-1} e^{-\frac{\beta^2}{q_{e,t}}}
    \EXP_t\left[1+\beta\left(g_{e,t}-\frac{\IND{e\in
    \bI_t}g_{e,t}}{q_{e,t}}\right)+\beta^2\left(g_{e,t}-\frac{\IND{e\in
    \bI_t}g_{e,t}}{q_{e,t}}\right)^2 \right]
    \label{lempr2}\\
&=& Z_{t-1} e^{-\frac{\beta^2}{q_{e,t}}}
    \EXP_t\left[1+\beta^2\left(g_{e,t}
      -\frac{\IND{e\in \bI_t}g_{e,t}}{q_{e,t}}\right)^2
          \right]
    \label{lempr3} \\
&\le& Z_{t-1} e^{-\frac{\beta^2}{q_{e,t}}}
    \EXP_t\left[1+\beta^2\left(\frac{\IND{e\in \bI_t}g_{e,t}}{q_{e,t}}
      \right)^2 \right] \nonumber \\
&\le& Z_{t-1} e^{-\frac{\beta^2}{q_{e,t}}}
        \left(1+\frac{\beta^2}{q_{e,t}}\right)
    \nonumber \\
&\le& Z_{t-1}. \label{lempr4}
\end{eqnarray}
Here (\ref{lempr2}) holds since $\beta \le 1$,
$g_{e,t}-\frac{\IND{e\in \bI_t}g_{e,t}}{q_{e,t}} \le 1$ and $e^x \le
1+x+x^2$ for $x \le 1$. (\ref{lempr3}) follows from $\EXP_t\left[
\frac{\IND{e\in \bI_t}g_{e,t}}{q_{e,t}}\right]=g_{e,t}$. Finally,
(\ref{lempr4}) holds by the inequality $1+x \le e^x$. Taking
expectations on both sides proves $\EXP[ Z_t ] \le \EXP[ Z_{t-1} ]$.
A similar argument shows that $\EXP[ Z_1 ]\le 1$, implying $\EXP[
Z_n ] \le 1$ as desired. \qed

\medskip
\noindent
{\bf Proof of Theorem~\ref{th:bandit}.}
As usual in the analysis of exponentially weighted average forecasters,
we start with
bounding the quantity $\ln \frac{\ol{W}_n}{\ol{W}_0}$.
On the one hand, we have the lower
bound
\begin{equation}
\label{eq:also} \ln\frac{\ol{W}_n}{\ol{W}_0}=\ln \sum_{\bi \in \cP}
e^{\eta G'_{\bi,n}} - \ln N \ge \eta \max_{\bi\in\cP}
G'_{\bi,n}-\ln N~.
\end{equation}

To derive a suitable upper bound, first notice that the condition
$\eta \le \frac{\gamma}{2K|\C|}$ implies $\eta g'_{\bi,t} \le 1$ for
all $\bi$ and $t$, since
\[
\eta g'_{\bi,t}=\eta \sum_{e \in \bi} g'_{e,t} \le
\eta \sum_{e\in\bi} \frac{1+\beta}{q_{e,t}}\le \frac{\eta K(1+\beta)|\C|}{\gamma}
\le 1
\]
where the second inequality follows because $q_{e,t} \ge
\gamma/|\C|$ for each $e\in E$.

\noindent Therefore, using the fact that $e^x \le 1+x+x^2$ for all
$x \le 1$, for all $t= 1,2,\ldots$ we have
\begin{eqnarray}
\ln \frac{\ol{W}_t}{\ol{W}_{t-1}}&=& \ln \sum_{\bi\in\cP} \frac{\bw_{\bi,t-1}}{\ol{W}_{t-1}}
e^{\eta g'_{\bi,t}} \nonumber \\*
&=& \ln \left(\sum_{\bi \in \cP} \frac{p_{\bi,t}
-\frac{\gamma}{|\C|}\IND{\bi \in \C}}{1-\gamma} e^{\eta g'_{\bi,t}} \right)
\label{pr1} \\
&\le & \ln \left(\sum_{\bi \in \cP} \frac{p_{\bi,t}
-\frac{\gamma}{|\C|}\IND{\bi \in \C}}{1-\gamma} \left(1+\eta g'_{\bi,t}
+\eta^2 {g'}^2_{\bi,t}\right) \right) \nonumber \\
&\le& \ln \left(1+\sum_{\bi \in \cP} \frac{p_{\bi,t}}{1-\gamma}
\left(\eta g'_{\bi,t} +\eta^2 {g'}^2_{\bi,t}\right) \right) \nonumber \\
&\le & \frac{\eta}{1-\gamma} \sum_{\bi\in\cP} p_{\bi,t} g'_{\bi,t} +
\frac{\eta^2}{1-\gamma} \sum_{\bi\in\cP} p_{\bi,t} {g'}^2_{\bi,t}
\label{eq:felso}
\end{eqnarray}
where (\ref{pr1}) follows form the definition of $p_{\bi,t}$, and
(\ref{eq:felso}) holds by the inequality $\ln(1+x)\le x$ for all $x>-1$.

Next we bound the sums in (\ref{eq:felso}). On the one hand,
\begin{eqnarray*}
\sum_{\bi\in\cP} p_{\bi,t} g'_{\bi,t} & = & \sum_{\bi\in\cP}
p_{\bi,t} \sum_{e\in\bi}g'_{e,t}=
\sum_{e\in E } g'_{e,t} \sum_{\bi\in\cP:e\in\bi}p_{\bi,t} \\
&=& \sum_{e\in E } g'_{e,t}q_{e,t} = g_{\bI_t,t}+| E |\beta.
\end{eqnarray*}
On the other hand,
\begin{eqnarray*}
\sum_{\bi\in\cP} p_{\bi,t} {g'}^2_{\bi,t}
&=& \sum_{\bi\in\cP} p_{\bi,t} \left(\sum_{e\in\bi}g'_{e,t}\right)^2 \\
&\le& \sum_{\bi\in\cP} p_{\bi,t} K\sum_{e\in\bi}{g'}^2_{e,t} \\
&=& K \sum_{e\in E }{g'}^2_{e,t} \sum_{\bi\in\cP:e\in\bi} p_{\bi,t} \\
&=& K \sum_{e\in E }{g'}^2_{e,t} q_{e,t} \\
&=&  K \sum_{e\in E } q_{e,t} g'_{e,t}\frac{\beta
+\IND{e \in \bI_t} g_{e,t}}{q_{e,t}} \\
&\le& K(1+\beta)\sum_{e\in E } g'_{e,t}
\end{eqnarray*}
where the first inequality is due to the inequality between the
arithmetic and quadratic mean, and the second one holds because
$g_{e,t} \le 1$.
Therefore,
\[
\ln \frac{\ol{W}_t}{\ol{W}_{t-1}} \le
\frac{\eta}{1-\gamma}\left(g_{\bI_t,t}+| E |\beta\right)
+\frac{\eta^2K(1+\beta)}{1-\gamma}\sum_{e\in E }g'_{e,t}~.
\]
\noindent Summing for $t=1,\ldots,n$, we obtain
\begin{eqnarray}
\ln \frac{\ol{W}_n}{\ol{W}_0} &\le&
\frac{\eta}{1-\gamma}\left(\wh{G}_n+n| E |\beta\right)
+\frac{\eta^2K(1+\beta)}{1-\gamma}\sum_{e\in E } G'_{e,n} \notag\\
&\le& \frac{\eta}{1-\gamma}\left(\wh{G}_n+n| E |\beta\right)
+\frac{\eta^2K(1+\beta)}{1-\gamma} |\C| \max_{\bi\in\cP} G'_{\bi,n} \label{eq:upperbound}
\end{eqnarray}
where the second inequality follows since $\sum_{e\in E }G'_{e,n}
\le \sum_{\bi\in\C} G'_{\bi,n}$. Combining the upper bound with the
lower bound (\ref{eq:also}), we obtain
\begin{equation}\label{eq:low+upper}
\wh{G}_n \ge \left(1-\gamma-\eta
K(1+\beta)|\C|\right)\max_{\bi\in\cP}G'_{\bi,n}
-\frac{1-\gamma}{\eta}\ln N - n| E |\beta.
\end{equation}

\noindent Now using Lemma \ref{le:supermartingale2} and applying the
union bound, for any $\delta \in (0,1)$ we have that, with probability
at least $1-\delta$,
\[ \wh{G}_n \ge \left(1-\gamma-\eta
K(1+\beta)|\C|\right)\left(\max_{\bi\in\cP}G_{\bi,n}
-\frac{K}{\beta}\ln\frac{| E
|}{\delta}\right)-\frac{1-\gamma}{\eta}\ln N - n| E |\beta~,
\]
where we used $1-\gamma-\eta K(1+\beta)|\C| \ge 0$ which follows from the
assumptions of the theorem.

\noindent Since $\wh{G}_n=Kn-\wh{L}_n$ and $G_{\bi,n}=Kn-L_{\bi,n}$ for all $\bi \in \cR$,
we have
\begin{eqnarray*}
\wh{L}_n &\le& Kn\left(\gamma+\eta(1+\beta)K|\C|\right)+
\left(1-\gamma-\eta(1+\beta)K|\C|\right)\min_{\bi\in\cP} L_{\bi,n} \nonumber\\
&&\mbox{}+\left(1-\gamma-\eta(1+\beta)K|\C|\right)\frac{K}{\beta}\ln\frac{|
E |}{\delta} +\frac{1-\gamma}{\eta}\ln N + n| E |\beta
\end{eqnarray*}
with probability at least $1-\delta$. This implies
\begin{eqnarray*}
\wh{L}_n - \min_{\bi\in\cP} L_{\bi,n} &\le&
Kn\gamma+\eta(1+\beta)nK^2|\C|
+\frac{K}{\beta}\ln\frac{| E |}{\delta}+\frac{1-\gamma}{\eta}\ln N + n| E |\beta \\
&\le& Kn\gamma+2\eta nK^2|\C| +\frac{K}{\beta}\ln\frac{| E
|}{\delta}+\frac{\ln N}{\eta} + n| E |\beta
\end{eqnarray*}
with probability at least $1-\delta$, which is the first statement of
the theorem. Setting
\[
\beta=\sqrt{\frac{K}{n| E |}\ln\frac{| E |}{\delta}} \qquad
\text{and} \qquad \gamma=2 \eta K|\C|
\]
results in the inequality
\[
\wh{L}_n - \min_{\bi\in\cP} L_{\bi,n} \le 4\eta nK^2 |\C|+\frac{\ln
N}{\eta}+2\sqrt{nK| E |\ln\frac{| E |}{\delta}}
\]
which holds with probability at least $1-\delta$ if $n \ge (K/| E
|)\ln (| E |/\delta)$ (to ensure $\beta \le 1$). Finally, setting
\[
\eta=\sqrt{\frac{\ln N}{4nK^2|\C|}}
\]
yields the last statement of the theorem  ($n \ge 4\ln N|\C|$ is
required to ensure $\gamma\le 1/2$). \qed

\medskip

Next we analyze the computational complexity of the algorithm.
The next result shows that the algorithm is feasible as its complexity
is linear in the size (number of edges) of the graph.

\begin{theorem}
\label{thmcomplex}
The proposed algorithm can be implemented efficiently with time complexity
$O(n|E|)$ and space complexity $O(|E|)$.
\end{theorem}
\proof The two complex steps of the algorithm are steps (a) and (b),
both of which can be computed, similarly to Takimoto and Warmuth
\cite{TaWa03}, using dynamic programming. To perform these steps
efficiently, first we order the vertices of the graph. Since we have
an acyclic directed graph, its vertices can be labeled (in $O(|E|)$ time)
from $1$ to $|V|$ such that $u=1$,  $v=|V|$, and  if $(v_1,v_2) \in
E$, then $v_1<v_2$.  For any pair of vertices $u_1<v_1$ let
$\cP_{u_1,v_1}$ denote the set of paths from $u_1$ to $v_1$, and for
any vertex $s \in V$, let
\[
H_t(s)=\sum_{\bi \in \cP_{s,v}} \prod_{e \in \bi} w_{e,t}
\]
and
\[
\widehat{H}_t(s)=\sum_{\bi \in \cP_{u,s}} \prod_{e \in \bi} w_{e,t}~.
\]
Given the edge weights $\{w_{e,t}\}$, $H_t(s)$ can be computed
recursively for $s=|V|-1,\ldots,1$, and $\widehat{H}_t(s)$ can be
computed recursively for $s=2,\ldots,|V|$ in $O(|E|)$ time (letting
$H_t(v)=\widehat{H}_t(u)=1$ by definition). In step (a), first one
has to decide with probability $\gamma$ whether $\bI_t$ is generated
according to the graph weights, or it is chosen uniformly from $\C$.
If $\bI_t$ is to be drawn according to the graph weights, it can be
shown that its vertices can be chosen one by one such that if the
first $k$ vertices of $\bI_t$ are $v_0=u,v_1,\ldots,v_{k-1}$, then
the next vertex of $\bI_t$ can be chosen to be any $v_k>v_{k-1}$,
satisfying $(v_{k-1},v_k)\in E$, with probability
$w_{(v_{k-1},v_k),t-1} H_{t-1}(v_{k})/H_{t-1}(v_{k-1})$.  The other
computationally demanding step, namely step (b), can be performed
easily by noting that for any edge $(v_1,v_2)$,
\begin{eqnarray*}
q_{(v_1,v_2),t}&=& (1-\gamma) \frac{\widehat{H}_{t-1}(v_1)
w_{(v_1,v_2),t-1} H_{t-1}(v_2)}{H_{t-1}(u)} \\ &&\;\; + \;\;
\gamma\frac{\left|\{\bi \in \C: (v_1,v_2)\in\bi\}\right|}{|\C|}
\end{eqnarray*}
as desired.\qed

\section{A combination of the label efficient and bandit settings}
\label{sec:leband}

In this section we investigate a combination of the multi-armed
bandit and the label efficient problems. This means that the decision
maker only has access to the loss of
the chosen path upon request and the total number of requests must be
bounded by a constant $m$.
This combination is motivated by some
applications, in which feedback information is costly to obtain.

In the general label efficient decision problem, after taking an
action, the decision maker has the option to query the losses of all
possible actions.  
For this
problem, Cesa-Bianchi {\em et al}.\  \cite{CeLuSt05} proved an upper
bound on the normalized regret of order $O(K\sqrt{\ln (4N/\delta)/(m)})$
which holds with probability at least $1-\delta$.

Our model of the label-efficient bandit problem for shortest paths is
motivated by an application to a particular packet switched network
model. This model, called the cognitive packet network, was introduced
by Gelenbe {\em et al}.\ \cite{GeGeLeLiSu04,GeLeXh01}. In these
networks a particular type of packets, called smart packets, are used
to explore the network (e.g., the delay of the chosen path). These
packets do not carry any useful data; they are merely used for
exploring the network.  The other type of packets are the data
packets, which do not collect any information about their paths.  The
task of the decision maker is to send packets from the source to the
destination over routes with minimum average transmission delay (or
packet loss). In this scenario, smart packets are used to query the
delay (or loss) of the chosen path. However, as these packets do not transport
information, there is a tradeoff between the number of queries and the
usage of the network. If data packets are on the average $\alpha$
times larger than smart packets (note that typically
$\alpha\gg 1$) and $\eps$ is the proportion of
time instances when smart packets are used to explore the network, 
then $\eps/(\eps+\alpha (1-\eps))$ is the proportion
of the bandwidth sacrificed for well informed routing decisions.

We study a combined algorithm which, at each time slot $t$, queries
the loss of the chosen path with probability $\eps$ (as in the 
solution of the label efficient problem proposed in \cite{CeLuSt05}), 
and, similarly to the multi-armed bandit case,
computes biased estimates $g'_{\bi,t}$ of the true gains
$g_{\bi,t}$.
Just as in the
previous section, it is assumed that each path of the graph is of
the same length $K$.

The algorithm differs from our bandit algorithm of the previous
section only in step (c), which is modified in the spirit of
\cite{CeLuSt05}. The modified step is given below:

\bookbox{
\begin{center}
MODIFIED STEP FOR THE LABEL EFFICIENT BANDIT ALGORITHM FOR SHORTEST
PATHS
\end{center} \label{calgo}
\begin{itemize}
\item[(c')] Draw a Bernoulli random variable $S_t$ with
$\PROB{(S_t=1)}=\eps$, and
compute the estimated gains 
\[
g'_{e,t}=\begin{cases}
\frac{g_{e,t}+\beta}{\eps q_{e,t}}S_t & \text{ if $e \in \bI_t$ } \\ 
\frac{\beta}{\eps q_{e,t}}S_t & \text{ if $e \notin \bI_t$ }~.
\end{cases}
\]
\end{itemize}
}

The performance of the algorithm is analyzed in the next theorem,
which can be viewed as a combination of Theorem \ref{th:bandit} in
the preceding section and Theorem~2 of \cite{CeLuSt05}.

\begin{theorem} \label{th:lebandit}
For any $\delta \in (0,1)$, $ \eps \in (0,1]$, parameters $\eta =
\sqrt{\frac{\eps \ln N}{4n K^2|\C|}}$, $\gamma = \frac{2\eta
K|\C|}{\eps} \le 1/2$, and
$\beta=\sqrt{\frac{K}{n|E|\eps}\ln\frac{2|E|}{\delta}}\le 1$, and for
all
\[
n \ge \frac{1}{\eps}\max \left\{\frac{K^2 \ln^2(2|E|/\delta)}{|E|\ln
N}, \frac{|E|\ln(2|E|/\delta)}{K}, 4|\C|\ln N\right\}
\]
the performance of the algorithm defined above can be bounded, with
probability at least $1-\delta$, as
\begin{eqnarray*}
\lefteqn{\frac{1}{n}\left(
\wh{L}_n-\min_{\bi\in\cP} \sum_{t=1}^n \ell_{\bi,t}\right)} \\
&\le& \sqrt{\frac{K}{n\eps}} \left( 4\sqrt{K|\C|\ln N } + 5
\sqrt{|E|\ln \frac{2|E|}{\delta}} + \sqrt{8K \ln
\frac{2}{\delta}}\right)
+ \frac{4K}{3n\eps}\ln \frac{2N}{\delta} \\
&\le& \frac{27K}{2} \sqrt{\frac{|E|\ln\frac{2 N}{\delta}}{n \eps}}~.
\end{eqnarray*}
\end{theorem}

If $\eps$ is chosen as $(m-\sqrt{2m\ln(1/\delta)})/n$ then, with
probability at least $1-\delta$, the total number of queries is
bounded by $m$ (see \cite[Lemma 6.1]{CeLu06}) and the performance bound
above is of the order of $K\sqrt{|E|\ln(N/\delta)/m}$.

Similarly to Theorem \ref{th:bandit}, we need a
lemma which reveals the connection between the true and the estimated
cumulative losses. However, here we need a more careful analysis because the
 ``shifting term''
$\frac{\beta}{\eps q_{e,t}}S_t$, is a random variable.

\begin{lemma}
\label{le:supermartingale3} For any $ 0 < \d <1$, $ 0 < \eps \le 1$,
for any
\[
n \ge \frac{1}{\eps}\max\left\{\frac{K^2 \ln^2(2|\E|/\d)}{|\E|\ln
N},\frac{K \ln(2|\E|/\d)}{|\E|} \right\}~,
\]
 parameters
\[
\frac{2\eta K|\C|}{\eps} \le \gamma, \qquad  \eta = \sqrt{\frac{\eps
\ln N }{4n K^2|\C| }} \quad \text{ and } \quad
\beta=\sqrt{\frac{K}{n|\E|\eps}\ln\frac{2|\E|}{\delta}}\le 1~,
\]
 and $\e \in \E$, we have

\[
\PROB \left[G_{\e,n}>G'_{\e,n}+\frac{4}{\b\eps}
\ln\frac{2|\E|}{\delta}\right]\le \frac{\d}{2|\E|}~.
\]
\end{lemma}
\proof Fix $\e\in\E$. Using (\ref{eq:chernoff}) with
 $u=\frac{4}{\b\eps} \ln\frac{2|\E|}{\delta}$ and $c=\frac{\b\eps}{4}$,
 it suffices to prove for all $n$ that
\[
\EXP \left[ e^{c(G_{\e,n}-G'_{\e,n})}\right]\le 1~. 
\]
Similarly to Lemma \ref{le:supermartingale2} we introduce
$Z_t=e^{c(G_{\e,t}-G'_{\e,t})}$ and
we show that $Z_1, \ldots, Z_n$ is a supermartingale, that is
$\EXP_t[ Z_t ]\le Z_{t-1}$ for $t \ge 2$ where $\EXP_t$ denotes
$\EXP[\cdot|(\bI_1,S_1),\ldots,(\bI_{t-1},S_{t-1})]$.
Taking conditional expectations, we obtain
\begin{eqnarray}
\EXP_t[Z_t] &=&Z_{t-1} 
\EXP_t\left[e^{c\left(g_{\e,t}-\frac{ \I_{\{\e\in I_t\}}S_t g_{\e,t}+
        S_t\b}{q_{\e,t}\eps}\right)}\right]
        \notag \\
&\le&Z_{t-1} \EXP_t\left[1
    +c\left(g_{\e,t}-\frac{\I_{\{\e\in I_t\}}S_tg_{\e,t}+S_t\b}{q_{\e,t}\eps}\right) \right. \notag\\
& &   \left.
+c^2\left(g_{\e,t}-\frac{\I_{\{\e\in
I_t\}}S_tg_{\e,t}+S_t\b}{q_{\e,t}\eps}\right)^2
    \right]. \label{1+ex}
\end{eqnarray}
Since
\[
    \EXP_t \left[g_{\e,t}-\frac{\I_{\{\e\in I_t\}}S_tg_{\e,t}+S_t\b}{q_{\e,t}\eps}
     \right]
    =-\frac{\b}{q_{\e,t}}
\]
and
\begin{align*}
    \EXP_t \left[\left(g_{\e,t}-\frac{\I_{\{\e\in I_t\}}S_tg_{\e,t}}{q_{\e,t}\eps}\right)^2\right]
    \le  \EXP_t\left[\left(\frac{\I_{\{\e\in I_t\}}S_tg_{\e,t}}{q_{\e,t}\eps}\right)^2\right]
    \le  \frac{1}{q_{e,t}\eps}
\end{align*}
we get from (\ref{1+ex}) that
\begin{eqnarray}
\lefteqn{\EXP_t[Z_t]} \notag\\
&\le & Z_{t-1} \EXP_t\left[1
    -\frac{c\b}{q_{e,t}}
    + \frac{c^2}{q_{e,t}\eps} 
   +c^2 \left( \frac{2\I_{\{\e\in I_t\}}S_t g_{\e,t}
\b}{q_{\e,t}^2\eps^2}
    -  \frac{2g_{\e,t} S_t\b}{q_{\e,t}\eps}
    + \frac{S_t \b^2}{q_{\e,t}^2 \eps^2} \right)
    \right]
    \notag\\
&\le&Z_{t-1} \left(1 + \frac{c}{q_{e,t}} \left(
    -\b + \frac{c}{\eps} + c\b \left(
    \frac{2}{\eps}+\frac{\b}{q_{e,t}\eps} \right)
    \right) \right).
    \label{norandom}
\end{eqnarray}

Since $c=\b \eps/4$ we have
\begin{eqnarray}
-\b + \frac{c}{\eps} + c\b
\left(\frac{2}{\eps}+\frac{\b}{q_{e,t}\eps}\right)
&=& -\frac{3\b}{4} + \frac{\b^2\eps}{4}\left(\frac{2}{\eps}+\frac{\b}{q_{e,t}\eps} \right) \notag\\*
&=& -\frac{3\b}{4} + \frac{\b^2}{2}+\frac{\b^3}{4q_{\e,t}} \notag\\
&\le& -\frac{\b}{4} + \frac{\b^3}{4q_{\e,t}} \notag \\
&\le& -\frac{\b}{4} + \frac{\b^3|\C|}{4\gamma} \label{pitlowerbound}\\
&\le& 0, \label{gammalowerbound}
\end{eqnarray}
where (\ref{pitlowerbound}) follows from $q_{\e,t} \ge \frac{\gamma}{|\C|}$ and (\ref{gammalowerbound})
holds by
\[
\frac{\b^2|\C|}{\gamma} \le \frac{\b^2\eps}{2\eta K} \le 1~,
\]
and the last inequality is ensured by $n \ge \frac{K^2
\ln^2(2|\E|/\d)}{ \eps |\E|\ln N}$, the assumption of the lemma.

 Combining (\ref{norandom}) and (\ref{gammalowerbound})
we get that $\EXP_t[Z_t] \le Z_{t-1}$. Taking expectations on
both sides of the inequality, we get $\EXP[Z_t] \le \EXP[Z_{t-1}]$ and
since $\EXP[Z_1] \le 1$, we obtain $\EXP[Z_n] \le 1$ as desired. \qed

\medskip
\noindent
{\bf Proof of Theorem~\ref{th:lebandit}.}
The proof of the theorem is a generalization of that of
Theorem~\ref{th:bandit}, and follows the same lines with some extra
technicalities to handle the effects of the modified step (c').
Therefore, in the following we emphasize only the differences.
First note that (\ref{eq:also}) and (\ref{eq:felso}) also hold in this case.
Bounding the sums in (\ref{eq:felso}), one obtains
\[
\sum_{\bi\in\cP} p_{\bi,t} g'_{\bi,t} = \frac{S_t}{\eps} \left(
g_{\bI_t,t}+|E|\beta \right) \]
and
\[
\sum_{\bi\in\cP} p_{\bi,t} {g'}^2_{\bi,t} \le
\frac{1}{\eps}K(1+\beta)\sum_{e\in E} g'_{e,t}~.
\]
Plugging these bounds into (\ref{eq:felso}) and summing for $t=1,\ldots,n$, we
obtain
\begin{eqnarray*}
\ln \frac{\ol{W}_n}{\ol{W}_0} 
&\le& \frac{\eta}{1-\gamma} \sum_{t=1}^n \frac{S_t}{\eps} \left( g_{I_t,t}+|\E|\b\ \right)
+\frac{\eta^2K(1+\b)}{(1-\gamma)\eps} |\C| \max_{\bi\in\cP} G'_{\bi,n}~.
\end{eqnarray*}
Combining the upper bound with the lower bound
(\ref{eq:also}), we obtain
\begin{eqnarray}
\sum_{t=1}^n \frac{S_t}{\eps} \left( g_{\bI_t,t}+|E|\beta\ \right)
&\ge& \left(1\!-\!\gamma\!-\!\frac{\eta
K(1+\beta)|\C|}{\eps}\right)\max_{\bi\in\cP} G'_{\bi,n}
\!-\!\frac{\ln N}{\eta}~.   \label{eq:combuplow}
\end{eqnarray}
To relate the left-hand side of the above inequality to the real
gain $\sum_{t=1}^n g_{\bI_t,t}$, notice that
\[
X_t = \frac{S_t}{\eps}\left(g_{\bI_t,t}+|E|\beta\right) -
\left(g_{\bI_t,t}+|E|\beta\right)
\]
is a martingale difference sequence with respect to $(\bI_1,S_1),
(\bI_2,S_2),\ldots$. Now for all $t=1,\ldots,n$, we have the bound
\begin{eqnarray}
\EXP \left[X_t^2|(\bI_1,S_1),\ldots,(\bI_{t-1},S_{t-1})\right]
&\le&\EXP \left[ \frac{S_t}{\eps^2}(g_{\bI_t,t}+|\E|\b)^2
  \bigg|(\bI_1,S_1),\ldots,(\bI_{t-1},S_{t-1})\right] \notag\\
&\leq& \frac{(K+|\E|\b)^2}{\eps} \notag\\
&\leq& \frac{4K^2}{\eps}\defeq\sigma^2, \label{eq:betabound}
\end{eqnarray}
where (\ref{eq:betabound}) holds by $n \ge \frac{|\E| \ln (2|\E|/\d)}{K\eps}$
(to ensure $\b|\E| \le K$). We know that
\[
X_t \in \left[-2K,\left(\frac{1}{\eps}-1\right)2K\right]
\]
 for all $t$. Now apply Bernstein's inequality
for martingale differences (see Lemma \ref{le:Bernstein} in the Appendix)
to obtain
\begin{equation}
\PROB \left[\sum_{t=1}^n X_t >u \right] \le \frac{\delta}{2}~, \label{generalbernstein}
\end{equation}
where
\[
u=\sqrt{2n \frac{4K^2}{\eps} \ln \left(\frac{2}{\delta}\right)} + \frac{4K}{3\eps}\ln \left(\frac{2}{\delta}\right).
\]
From (\ref{generalbernstein}) we get
\begin{equation}
\PROB \left[\sum_{t=1}^n \frac{S_t}{\eps}\left(g_{\bI_t,t}+|\E|\b\right)  \ge
\wh{G}_n + \b n|\E| + u \right] \le \frac{\delta}{2}~.
 \label{eq:deltabernstein}
\end{equation}

Now Lemma \ref{le:supermartingale3}, the union bound,  and
(\ref{eq:deltabernstein}) combined with (\ref{eq:combuplow}) yield, with
probability at least $1-\delta$,
\begin{eqnarray*}
\wh{G}_n & \ge& \left( 1-\gamma-\frac{\eta
K(1+\beta)|\C|}{\eps} \right)
\left( \max_{\bi\in\cP} G_{\bi,n} - \frac{4K}{\beta \eps} \ln \frac{2|E|}{\delta} \right) \\
&&-\frac{\ln N}{\eta} - \beta
n|E|-u
\end{eqnarray*}
since the coefficient of $G'_{i,n}$ is greater than zero
by the assumptions of the theorem.

Since $\wh{G}_n=Kn-\wh{L}_n$ and $G_{\bi,n}=Kn-L_{\bi,n}$, we have
\begin{eqnarray*}
\wh{L}_n &\le& \left(1-\gamma-\frac{K(1+\b)\eta|\C|}{\eps}\right)\min_{\bi\in\cP} L_{\bi,n}
+Kn\left(\gamma+\frac{K(1+\b)\eta|\C|}{\eps}\right)\\
& &
+\left(1-\gamma-\frac{K(1+\b)\eta|\C|}{\eps}\right) \frac{4K}{\beta \eps} \ln \frac{2|E|}{\delta}
+ \b n |\E|
+\frac{\ln N}{\eta}
+u\\
&\le & \min_{\bi\in\cP} L_{\bi,n}
+ Kn\left(\gamma+\frac{K(1+\b)\eta|\C|}{\eps}\right)
+ 5 \b n |\E|
+\frac{\ln N}{\eta}
+u~,
\end{eqnarray*}
where we used the fact that
$\frac{K}{\beta \eps} \ln \frac{2|E|}{\d} = \b n |\E|$.

Substituting the value of $\b$, $\eta$ and  $\gamma$, we have
\begin{align*}
\wh{L}_n-\min_{\bi\in\cP} L_{\bi,n}
\le&
Kn \frac{2K\eta|\C|}{\eps} +Kn \frac{2K\eta|\C|}{\eps}
+\frac{\ln N}{\eta}
+ 5 \b n |\E|
+ u\\
\le&
4K \sqrt{\frac{n |\C|\ln N}{\eps}}
+ 5 \sqrt{\frac{n|\E|K \ln (2|\E|/\d)}{\eps}}
+ u\\
\le&
\sqrt{\frac{nK}{\eps}}
\left(
4\sqrt{K|\C|\ln N }
+ 5 \sqrt{|\E|\ln (2|\E|/\d)}
+ \sqrt{8K \ln \left(2/\delta\right)}\right)\\
& + \frac{4K}{3\eps}\ln \left(2/\delta\right)
\end{align*}
as desired.\qed

\section{A bandit algorithm for tracking the shortest path}
\label{sec:tracking}

Our goal in this section is to extend the
bandit algorithm  so that it is able to compete with time-varying paths under
small computational complexity.
This is a variant of the problem known as \emph{tracking the best expert};
see, for example,
Herbster and Warmuth \cite{HeWa98},
Vovk~\cite{Vov99},
Auer and Warmuth \cite{AW98},
Bousquet and Warmuth \cite{BoWa02},
Herbster and Warmuth \cite{HeWa01}.

To describe the loss the decision maker is compared to, consider the
following ``$m$-partition'' prediction scheme: the sequence of paths
is partitioned into $m+1$ contiguous segments, and on each segment the
scheme assigns exactly one of the $N$ paths. Formally, an
$m$-partition $\P(n,m,\bt,\ui)$ of the $n$ paths is given by an
$m$-tuple $\bt=(t_1,\ldots,t_m)$ such that
$t_0=1<t_1<\cdots<t_m<n+1=t_{m+1}$, and an $(m+1)$-vector
$\ui=(\bi_0,\ldots,\bi_m)$ where $\bi_j \in \cP$. At each time instant
$t$, $t_j\le t < t_{j+1}$, path $\bi_j$ is used to predict the best
path. The cumulative loss of a partition $\P(n,m,\bt,\ui)$ is
\[
L(\P(n,m,\bt,\ui))=\sum_{j=0}^m \sum_{t=t_j}^{t_{j+1}-1}
\ell_{\bi_j,t}= \sum_{j=0}^m \sum_{t=t_j}^{t_{j+1}-1} \sum_{e \in
\bi_j} \ell_{e,t}.
\]

The goal of the decision maker is to perform as well as the best
time-varying path (partition), that is, to keep
the normalized
regret
\[
\frac{1}{n}\left(\wh{L}_n-\min_{\bt,\ui} L(\P(n,m,\bt,\ui))\right)
\]
as small as possible (with high probability) for all possible
outcome sequences.

In the ``classical'' tracking problem there is a relatively
small number of ``base'' experts and the goal of the
decision maker is to predict as well as the best ``compound'' expert
(i.e., time-varying expert). However in our case, base experts
correspond to all paths of the graph between source and destination
whose number is typically exponentially large in the number of
edges, and therefore we cannot directly apply the
computationally efficient methods for tracking the best expert.
Gy\"orgy, Linder, and Lugosi \cite{GyLiLu05a} develop efficient
algorithms for tracking the best expert for certain large and
structured classes of base experts, including the shortest path
problem. The purpose of the following algorithm is to extend the methods
of \cite{GyLiLu05a}
to the bandit setting when the forecaster only observes the losses
of the edges on the chosen path.

\bookbox{ \label{alg:tracking}
\begin{center}
A BANDIT ALGORITHM FOR TRACKING SHORTEST PATHS
\end{center}
\textbf{Parameters:} real numbers $\beta>0$, $0 < \eta,\gamma < 1$,
$0\le \alpha \le 1$.
\\
\textbf{Initialization:} Set $w_{e,0}=1$ for each $e \in E$,
$\bw_{\bi,0}=1$ for each $\bi \in \cP$, and $\ol{W}_0=N$. For
each round $t=1,2,\ldots$
\begin{itemize}
\item[(a)] Choose a path $\bI_t$ according to the distribution $\bp_t$
defined by
\[
p_{\bi,t}=\begin{cases} (1-\gamma)\frac{\bw_{\bi,t-1}}{\ol{W}_{t-1}}
+ \frac{\gamma}{|\C|}
& \text{ if $\bi \in \C$;} \\
(1-\gamma)\frac{\bw_{\bi,t-1}}{\ol{W}_{t-1}} & \text{ if $\bi
\not\in \C$.}
\end{cases}
\]
\item[(b)] Compute the probability of choosing each edge $e$ as
\[
q_{e,t}=\sum_{\bi:e\in\bi}
p_{\bi,t}=(1-\gamma)\frac{\sum_{\bi:e\in\bi}
\bw_{\bi,{t-1}}}{\ol{W}_{t-1}}+ \gamma\frac{\left|\{\bi \in
\C:e\in\bi\}\right|}{|\C|}.
\]
\item[(c)] Calculate the estimated gains
\[
g'_{e,t}=\begin{cases}
\frac{g_{e,t}+\beta}{q_{e,t}} & \text{ if $e \in \bI_t$;} \\
\frac{\beta}{q_{e,t}} & \text{ otherwise.}
\end{cases}
\]
\item[(d)] Compute the updated weights
\begin{eqnarray*}
\bv_{\bi,t}&=&\bw_{\bi,t-1} e^{\eta g'_{\bi,t}} \\ 
\bw_{\bi,t}&=&(1-\alpha) \bv_{\bi,t} + \frac{\alpha}{N} \ol{W}_t
\end{eqnarray*}
where $g'_{\bi,t}=\sum_{e\in\bi}g'_{e,t}$ and $\ol{W}_t$ is the sum
of the total weights of the paths, that is,
\[
 \ol{W}_t=\sum_{\bi\in \cP} \bv_{\bi,t}=\sum_{\bi \in \cP} \bw_{\bi,t}.
\]
\end{itemize}
}

The following performance bounds shows that
the normalized regret with respect to
the best time-varying path which is allowed
to switch paths $m$ times is roughly of the order of
$K\sqrt{(m/n) |\C|\ln N}$.

\begin{theorem}\label{th:tracking}
For any $\delta \in (0,1)$ and parameters $0\le\gamma<1/2$,
$\alpha,\beta\in [0,1]$, and $\eta>0$ satisfying $2\eta
K|\C|\le \gamma$, the performance of the algorithm defined above
can be bounded, with probability at least $1-\delta$, as
\begin{eqnarray*}
\lefteqn{
\wh{L}_n - \min_{\bt,\ui}  L(\P(n,m,\bt,\ui))
 } \\*
&\le&
Kn\left(\gamma+\eta(1+\beta)K|\C|\right) +
\frac{K(m+1)}{\beta}\ln\frac{|E|(m+1)}{\delta}\\
&&\mbox{}+\beta n| E | +
\frac{1}{\eta}\ln\left(\frac{N^{m+1}}{\alpha^m(1-\alpha)^{n-m-1}
}\right)~.
\end{eqnarray*}
In particular, choosing \[
\beta=\sqrt{\frac{K(m+1)}{n|E |}\ln\frac{|\E
|(m+1)}{\delta}}, \qquad \gamma=2 \eta K |\C|, \qquad \alpha=\frac{m}{n-1},
\]
and
\[
\eta=\sqrt{\frac{(m+1)\ln N+m\ln\frac{e(n-1)}{m}}{4nK^2|\C|}}
\]
we have, for all $n\ge \max\left\{\frac{K(m+1)}{|E |}\ln\frac{|E
|(m+1)}{\delta}, 4|\C|D\right\}$,
\[
\wh{L}_n - \min_{\bt,\ui}  L(\P(n,m,\bt,\ui)) \le
2\sqrt{nK}\left(\sqrt{4K|\C| D}+ \sqrt{|E |(m+1)\ln\frac{|E
|(m+1)}{\delta}}\right),
\]
where
\[
D = (m+1) \ln N+m\left(1+\ln\frac{n-1}{m}\right).
\]
\end{theorem}

The proof of the theorem is a combination of that of our Theorem
\ref{th:bandit} and Theorem 1 of \cite{GyLiLu05a}.  We will need the
following three lemmas.

\begin{lemma}
\label{lempred2} For any $1\le t \le t' \le n$ and any $\bi \in \cP$,
\[
\frac{\bv_{\bi,t'}}{\bw_{\bi,{t-1}}} \ge e^{\eta G'_{\bi,[t,t']}}
(1-\alpha)^{t'-t}
\]
where $G'_{\bi,[t,t']} = \sum_{\tau=t}^{t'} g'_{\bi,\tau}$.
\end{lemma}

\proof The proof is a straightforward modification of the one in
Herbster and Warmuth \cite{HeWa98}.
From the definitions of $v_{\bi,t}$ and $w_{\bi, t}$
(see step (d) of the algorithm) it is clear that for any $\tau \ge
1$,
\[
\bw_{\bi,\tau} = (1-\alpha) \bv_{\bi,\tau} + \frac{\alpha}{N}
\ol{W}_\tau \ge (1-\alpha) e^{\eta g'_{\bi,\tau}}\bw_{\bi,\tau-1}~.
\]
Applying this equation iteratively for $\tau=t,t+1,\ldots,t'-1$, and
the definition of $\bv_{\bi,t}$ (step (d)) for $\tau=t'$, we obtain
\begin{align*}
\bv_{\bi,t'} = \bw_{\bi,t'-1} e^{\eta g'_{\bi,t'}} &\ge
 e^{\eta g'_{\bi,t'}}\prod_{\tau=t}^{t'-1} \left( (1-\alpha)e^{\eta
 g'_{\bi,\tau}} \right)\bw_{\bi,t-1}\\
 &= e^{\eta G'_{\bi,[t,t']}} (1-\alpha)^{t'-t}\bw_{\bi,t-1}
\end{align*}
which implies the statement of the lemma. \qed

\begin{lemma}
\label{lempred3} For any $t\ge 1$ and $\bi,\bj \in \cP$, we have
\[
\frac{\bw_{\bi,t}}{\bv_{\bj,t}} \ge \frac{\alpha}{N}
\]
\end{lemma}

\proof By the definition of $\bw_{\bi,t}$ we have
\[
\bw_{\bi,t}= (1-\alpha) \bv_{\bi,t} + \frac{\alpha}{N} \ol{W}_t
\ge \frac{\alpha}{N} \ol{W}_t \ge \frac{\alpha}{N}
\bv_{\bj,t}~.
\]
This completes the proof of the lemma. \qed

The next lemma is a simple corollary of Lemma \ref{le:supermartingale2}.
\begin{lemma}
\label{le:Gboundtracking} For any $\delta \in (0,1)$, $0 \le \beta \le
1$, $t \ge 1$ and $e \in E $ we have
\[
\PROB\left[G_{e,t}>G'_{e,t} + \frac{1}{\beta} \ln
\frac{|E|(m+1)}{\delta}\right] \le \frac{\delta}{|E|(m+1)}~.
\]
\end{lemma}

\proofof{Theorem~\ref{th:tracking}} The theorem is proved the same
way as Theorem \ref{th:bandit} until
(\ref{eq:upperbound}), that is,
\begin{eqnarray}
\ln \frac{\ol{W}_n}{\ol{W}_0} &\le&
\frac{\eta}{1-\gamma}\left(\wh{G}_n+n| E |\beta\right)
+\frac{\eta^2K(1+\beta)}{1-\gamma} |\C| \max_{\bi\in\cP}
G'_{\bi,n}\label{eq:trackingfelso}~.
\end{eqnarray}

Let $\P(n,m,\bt,\ui)$ be an arbitrary partition. Then the lower
bound is obtained as
\begin{equation}
\label{trackingalso} \ln\frac{\ol{W}_n}{\ol{W}_0} = \ln \sum_{\bj
\in \cP} \frac{\bw_{\bj,n}}{N} = \ln \sum_{\bj \in \cP}
\frac{\bv_{\bj,n}}{N} \ge \ln \frac{\bv_{\bi_m,n}}{N}
\end{equation}
(recall that $\bi_m$ denotes the path used in the last segment of
the partition). Now $v_{\bi_m,n}$ can be rewritten in the form of
the following telescoping product
\[
\bv_{\bi_m,n} =
\bw_{\bi_0,t_0-1}\frac{\bv_{\bi_0,t_1-1}}{\bw_{\bi_0,t_0-1}}
\prod_{j=1}^m
\left(\frac{\bw_{\bi_{j},t_j-1}}{\bv_{\bi_{j-1},t_j-1}}
\frac{\bv_{\bi_j,t_{j+1}-1}}{\bw_{\bi_{j},t_j-1}}\right).
\]
Therefore, applying Lemmas~\ref{lempred2} and~\ref{lempred3}, we
have
\begin{eqnarray*}
\bv_{\bi_m,n} &\ge& \bw_{\bi_0,t_0-1}
\left(\frac{\alpha}{N}\right)^m
\prod_{j=0}^m \left( (1-\alpha)^{t_{j+1}-1-t_j} e^{\eta G'_{\bi_j,[t_{j},t_{j+1}-1]}} \right)\\
&=& \left(\frac{\alpha}{N}\right)^m e^{\eta G'(\P(n,m,\bt,\ui))}
(1-\alpha)^{n-m-1} .
\end{eqnarray*}
Combining the lower bound with the upper bound
(\ref{eq:trackingfelso}), we have
\begin{eqnarray*}
\lefteqn{\ln\left(\frac{\alpha^m(1-\alpha)^{n-m-1}
}{N^{m+1}}\right) +
 \max_{\bt,\ui} \eta G'(\P(n,m,\bt,\ui))}\\
 &\le \frac{\eta}{1-\gamma}\left(\wh{G}_n+n| E |\beta\right) +
  \frac{\eta^2K(1+\beta)}{1-\gamma} |\C| \max_{\bi\in\cP} G'_{\bi,n}~,
\end{eqnarray*}
where we used the fact that $\P(n,m,\bt,\ui)$ is an arbitrary partition.
After rearranging and using $\max_{\bi\in\cP} G'_{\bi,n} \le
\max_{\bt,\ui}  G'(\P(n,m,\bt,\ui))$ we get
\begin{eqnarray*}
\wh{G}_n &\ge& \left(1-\gamma- \eta K(1+\beta) |\C|\right)
\max_{\bt,\ui}  G'(\P(n,m,\bt,\ui)) \\
&&-n| E |\beta -
\frac{1-\gamma}{\eta}\ln\left(\frac{N^{m+1}}{\alpha^m(1-\alpha)^{n-m-1}
}\right).
\end{eqnarray*}
Now since $1-\gamma-\eta K(1+\beta)|\C|\ge 0$, by the assumptions of
the theorem and from Lemma~\ref{le:Gboundtracking} with an application of
the union bound we obtain that, with
probability at least $1-\delta$,
\begin{align*}
\wh{G}_n \ge& \left(1-\gamma-\eta
K(1+\beta)|\C|\right)\left(\max_{\bt,\ui}  G(\P(n,m,\bt,\ui))
-\frac{K(m+1)}{\beta}\ln\frac{|E|(m+1)}{\delta}\right)\\
 &- n| E |\beta - \frac{1-\gamma}{\eta}\ln\left(\frac{N^{m+1}}{\alpha^m(1-\alpha)^{n-m-1} }\right).
\end{align*}
Since $\wh{G}_n=Kn-\wh{L}_n$ and
$G(\P(n,m,\bt,\ui))=Kn-L(\P(n,m,\bt,\ui))$, we have
\begin{eqnarray*}
\wh{L}_n &\le& \left(1-\gamma-\eta
K(1+\beta)|\C|\right)\min_{\bt,\bi}  L(\P(n,m,\bt,\ui))
+Kn\left(\gamma+\eta(1+\beta)K|\C|\right) \nonumber\\
&&\mbox{}+\left(1-\gamma-\eta(1+\beta)K|\C|\right)\frac{K(m+1)}{\beta}\ln\frac{|E|(m+1)}{\delta}
+ n| E |\beta \nonumber\\
&&\mbox{} +
\frac{1}{\eta}\ln\left(\frac{N^{m+1}}{\alpha^m(1-\alpha)^{n-m-1}
}\right).
\end{eqnarray*}
This implies that, with probability at least $1-\delta$,
\begin{eqnarray}
\lefteqn{
\wh{L}_n - \min_{\bt,\ui}  L(\P(n,m,\bt,\ui))
} \nonumber \\*
&\le&
Kn\left(\gamma+\eta(1+\beta)K|\C|\right) +
\frac{K(m+1)}{\beta}\ln\frac{|E|(m+1)}{\delta} \nonumber \\*
&&\mbox{}+n| E |\beta +
\frac{1}{\eta}\ln\left(\frac{N^{m+1}}{\alpha^m(1-\alpha)^{n-m-1}
}\right). \label{trackingsta}
\end{eqnarray}

To prove the second statement,
let $H(p) = -p\ln p - (1-p) \ln (1-p)$ and $D(p\parallel q) = p \ln
\frac{p}{q} + (1-p) \ln \frac{1-p}{1-q}$. Optimizing the value of
$\alpha$ in the last term of (\ref{trackingsta}) gives
\begin{eqnarray*}
\lefteqn{
\frac{1}{\eta}\ln\left(\frac{N^{m+1}}{\alpha^m(1-\alpha)^{n-m-1}
}\right)
} \\*
&=&\frac{1}{\eta}\left((m+1)\ln\left(N\right) + m \ln
\frac{1}{\alpha}
 + (n-m-1) \ln \frac{1}{1-\alpha}\right)\\
&=&\frac{1}{\eta}\big((m+1)\ln\left(N\right) + (n-1)(D_b(\alpha^*
\parallel \alpha)
 + H_b(\alpha^*))\big)
\end{eqnarray*}
where $\alpha^*= \frac{m}{n-1}$. For $\alpha=\alpha^*$ we obtain
\begin{eqnarray*}
\lefteqn{\frac{1}{\eta}\ln\left(\frac{N^{m+1}}{\alpha^m(1-\alpha)^{n-m-1}}\right)}
\\
 &=& \frac{1}{\eta}\left((m+1)\ln\left(N\right) +
(n-1)(H_b(\alpha^*))\right)\\
&=&\frac{1}{\eta}\left((m+1)\ln\left(N\right) + m \ln((n-1)/m) \right.\\
&& \left.+ (n-m-1) \ln (1+ m/(n-m-1))\right)\\
&\le&\frac{1}{\eta}\left((m+1)\ln\left(N\right) + m \ln((n-1)/m) + m
\right)  \\
&=&\frac{1}{\eta}\left((m+1)\ln\left(N\right) + m \ln \frac{e(n-1)}{m}\right)
\defeq \frac{1}{\eta}D
\end{eqnarray*}
where the inequality follows since $\ln(1+x)\le x$ for $x>0$. Therefore
\begin{eqnarray*}
\lefteqn{
\wh{L}_n - \min_{\bt,\ui}  L(\P(n,m,\bt,\ui))  } \\*
&\le&
Kn\left(\gamma+\eta(1+\beta)K|\C|\right) +
\frac{K(m+1)}{\beta}\ln\frac{|E|(m+1)}{\delta}
\mbox{}+n| E |\beta + \frac{1}{\eta}D~.
\end{eqnarray*}
which is the first statement of the theorem. Setting
\[
\beta=\sqrt{\frac{K(m+1)}{n|E |}\ln\frac{|E |(m+1)}{\delta}} \text{,
} \gamma=2 \eta K |\C| \text{, and } \eta=\sqrt{\frac{D}{4nK^2|\C|}}
\]
results in  the second statement of the theorem, that is,
\begin{eqnarray*}
\lefteqn{\wh{L}_n - \min_{\bt,\ui}  L(\P(n,m,\bt,\ui))}\hspace{1cm} \\
& \le &
2\sqrt{nK}\left(\sqrt{4K|\C| D}+ \sqrt{|E |(m+1)\ln\frac{|E
|(m+1)}{\delta}}\right). \QED
\end{eqnarray*}

Similarly to \cite{GyLiLu05a}, the proposed algorithm
has an alternative version, which is efficiently computable:

\bookbox{
\begin{center}
AN ALTERNATIVE BANDIT ALGORITHM FOR TRACKING SHORTEST PATHS
\end{center}
\label{alg:alternative} For $t=1$, choose $\bI_1$ uniformly from the
set $\cP$. For $t \ge 2$, 
\begin{itemize}
\item[(a)] Draw a Bernoulli random variable $\Gamma_t$ with
$\PROB(\Gamma_t=1)=\gamma$.
\item[(b)] 
If $\Gamma_t=1$, then choose $\bI_t$ uniformly from $\cal{C}$.
\item[(c)] If $\Gamma_t=0$,
\begin{itemize}
\item[(c1)] choose $\tau_t$ randomly according to the distribution
\begin{equation*}
\label{tau} \PROB\{\tau_t=t'\}=\begin{cases}
\frac{(1-\alpha)^{t-1} Z_{1,t-1}}{W_{t-1}} & \text{for $t'=1$} \\
\frac{\alpha(1-\alpha)^{t-t'}W_{t'}Z_{t',t-1}}{N W_t} & \text{for
$t'=2,\ldots,t$}
\end{cases}
\end{equation*}
where $Z_{t',t-1}= \sum_{\bi\in \cP} e^{\eta G'_{\bi,[t',t-1]}}$ for
$t'=1,\ldots,t-1$, and   $Z_{t,t-1}=N$;
\item[(c2)] given $\tau_t=t'$, choose $\bI_t$ randomly
according to the probabilities
\begin{equation*}
\label{hxmod} \PROB\{I_t= \bi|\tau_t=t'\}=\begin{cases}
\frac{e^{\eta G'_{\bi,[t',t-1]}}}{Z_{t',t-1}}  
& \text{for $t'=1,\ldots,t-1$} \\
\frac{1}{N} & \text{for $t'=t$}. \end{cases}
\end{equation*}
\end{itemize}
\end{itemize}
}

In a way completely analogous to \cite{GyLiLu05a}, in this alternative
formulation of the algorithm one can compute the probabilities
$\PROB\{\bI_t= \bi|\tau_t=t'\}$ and the normalization factors
$Z_{t',t-1}$ efficiently. Using the fact that the baseline bandit
algorithm for shortest paths has an $O(n|E|)$ time complexity by
Theorem~\ref{thmcomplex}, it follows from Theorem~3 of
\cite{GyLiLu05a} that the time complexity of the alternative bandit
algorithm for tracking the shortest path is $O(n^2|E|)$.

\section{An algorithm for the restricted multi-armed bandit problem}
\label{sec:generalizedbandit}

In this section we consider the situation where the decision maker
receives information only about the performance of the whole chosen path,
but the individual edge losses are not available. That is, the
forecaster has access to the sum $\ell_{\bI_t,t}$ of losses over
the chosen path $\bI_t$ but not to the losses
$\{\ell_{e,t}\}_{e\in\bI_t}$
 of the edges belonging to $\bI_t$.

This is the problem formulation considered by McMahan and Blum
\cite{McBl04} and Awerbuch and Kleinberg \cite{AwKl04}. 
McMahan and Blum provided a relatively simple 
algorithm whose regret is at most of the order of $n^{-1/4}$, while
Awerbuch and Kleinberg gave a more complex algorithm to achieve
$O(n^{-1/3})$ regret.
In this section we combine the strengths of these papers, and propose 
a simple algorithm with regret at most of the order of $n^{-1/3}$.
Moreover, our bound holds with high probability, while the
above-mentioned papers prove bounds for the expected regret only.  The
proposed algorithm uses ideas very similar to those of McMahan and Blum
\cite{McBl04}. The algorithm alternates between choosing a path from a
``basis'' $\bB$ to obtain  unbiased estimates of the loss (exploration
step), and choosing a path according to exponential weighting
based on these estimates.

A simple way to describe a path $\bi\in\cP$ is a binary row vector
with $|E|$ components which are indexed by the edges of the graph such
that, for each $e\in E$, the $e$th entry of the vector is $1$ if
$e\in\bi$ and $0$ otherwise. With a slight abuse of notation we will
also denote by $\bi$ the binary row vector representing path $\bi$.
In the previous sections, where the loss of each edge along the chosen
path is available to the decision maker, the complexity stemming from
the large number of paths was  reduced by representing all information
in terms of the edges, as the set of edges spans the
set of paths. That is, the vector corresponding to  a given path can be
expressed as the linear combination of the unit vectors associated
with the edges (the $e$th component of the unit vector representing
edge $e$ is $1$, while the other components are $0$).  While the
losses corresponding to such a spanning set are not observable in the
restricted setting of this section, one can choose a subset of $\cP$
that forms a \emph{basis}, that is, a collection of $b$ paths which
are linearly independent and each path in $\cP$ can be expressed as a
linear combination of the paths in the basis. We denote by $\bB$ the
$b \times |\E|$ matrix whose rows $\bb^1,\ldots,\bb^b$ represent the
paths in the basis. Note that $b$ is equal to the maximum number of linearly
independent vectors in $\{\bi:\bi \in \cP\}$, so $b\le
|E|$. 

Let $\bl^{(\E)}_t$ denote the (column) vector of the edge losses
$\{\ell_{e,t}\}_{e\in\E}$ at time $t$, and let
$\bl^{(\bB)}_t=(\ell_{\bb^1,t},\ldots,\ell_{\bb^b,t})^T$ be a
$b$-dimensional column vector whose components are the losses of the
paths in the basis $\bB$ at time $t$. If  $\alpha^{(\bi,\bB)}_{\bb^1},\ldots,
\alpha^{(\bi,\bB)}_{\bb^b}$ are the coefficients 
in the linear combination of the basis paths expressing path $\bi\in \cP$,
that is, $\bi = \sum_{j=1}^b  \alpha_{\bb^j}^{(\bi,\bB)} \bb^j$,
then the loss of  path $\bi \in \cP$  at time $t$ is given by 
\begin{equation}
\label{eq:pathestimate1}
\ell_{\bi,t}= 
\langle \bi,\bl^{(\E)}_t\rangle =    \sum_{j=1}^b
\alpha_{\bb^j}^{(\bi,\bB)} \langle\bb^j, \bl^{(\E)}_t  \rangle =
 \sum_{j=1}^b \alpha_{\bb^j}^{(\bi,\bB)} \ell_{\bb^j,t}
\end{equation}
where $\langle\cdot,\cdot\rangle$ denotes the standard inner product
in $\R^{|E|}$.
In the algorithm we obtain estimates $ \tilde{\ell}_{\bb^j,t}$ of
the losses of the basis paths and use (\ref{eq:pathestimate1}) to
estimate the loss of any $\bi\in \cP$ as
\begin{equation}
\label{eq:pathestimate}
\tilde{\ell}_{\bi,t}
= \sum_{j=1}^b \alpha_{\bb^j}^{(\bi,\bB)} \tilde{\ell}_{\bb^j,t}~.
\end{equation}

It is algorithmically advantageous to calculate the estimated
path losses $\tilde{\ell}_{\bi,t}$ from an intermediate estimate of
the individual edge losses. Let $\bB^+$ denote the the Moore-Penrose
inverse of $\bB$ defined by $\bB^+=\bB^T(\bB\bB^T)^{-1}$, where
$\bB^T$ denotes the transpose of $\bB$  and $\bB\bB^T$ is invertible
since the rows of $\bB$ are linearly independent. (Note that $\bB^+ =
\bB^{-1}$ if $b=|E|$).  Then letting
$\tilde{\bl}^{(\bB)}_t=
(\tilde{\ell}_{\bb^1,t},\ldots,\tilde{\ell}_{\bb^b,t})^T$ and
\[
\tilde{\bl}^{(\E)}_t=\bB^+\tilde{\bl}^{(\bB)}_t
\]
it is easy to see that $\tilde{\ell}_{\bi,t}$ in
(\ref{eq:pathestimate}) can be obtained as $\tilde{\ell}_{\bi,t} =
\langle \bi , \tilde{\bl}^{(\E)}_t\rangle $, or equivalently
\[
\tilde{\ell}_{\bi,t} = \sum_{e\in \bi} \tilde{\ell}_{e,t}.
\]
This form of the path losses allows for an efficient implementation of
exponential weighting via dynamic programming \cite{TaWa03}.
 
To analyze the algorithm we need an upper bound on the magnitude of
the coefficients $ \alpha_{\bb^j}^{(\bi,\bB)}$.  For this, we invoke
the definition of a barycentric spanner from \cite{AwKl04}: the basis
$\bB$ is called a \emph{$C$-barycentric spanner} if
$|\alpha_{\bb^j}^{(\bi,\bB)}| \le C$ for all $\bi \in \cP$ and
$j=1,\ldots,b$. Awerbuch and Kleinberg \cite{AwKl04} show that a
$1$-barycentric spanner exists if $\bB$ is a square matrix (i.e.,
$b=|\E|$) and give a low-complexity algorithm which finds a
$C$-barycentric spanner for $C>1$. We use their technique to show that
a 1-barycentric spanner also exists in case of a non-square $\bB$,
when the basis is chosen to maximize the absolute value of the
determinant of $\bB\bB^T$. As before, $b$ denotes the maximum number
of linearly independent vectors (paths) in $\cP$.

\begin{lemma}
\label{le:bary}
For a directed acyclic graph, the set of paths $\cP$ between two
dedicated nodes has a $1$-barycentric spanner. Moreover, let $\bB$ be
a $b\times |E|$ matrix with rows from $\cP$ such that $\det[\bB
\bB^T]\neq 0$. If $\bB_{-j,\bi}$ is the matrix obtained from $\bB$ by
replacing its $j$th row by $\bi \in \cP$ and
\begin{equation}
\label{eq:cspanner}
\left|\det\left[\bB_{-j,\bi}\bB_{-j,\bi}^T\right]\right|
\le C^2 \left|\det\left[\bB\bB^T\right]\right|
\end{equation}
for all $j=1,\ldots,b$ and $\bi \in \cP$, then
$\bB$ is a $C$-barycentric spanner. 
\end{lemma}
\proof Let $\bB$ be a basis of $\cP$ with rows $\bb^1,\ldots,\bb^b\in
\cP$ that maximizes $|\det[\bB\bB^T]|$. Then, for any path $\bi \in
\cP$, we have $\bi = \sum_{j=1}^b \alpha_{\bb^j}^{(\bi,\bB)} \bb^j$
for some coefficients $\{\alpha^{(\bi,\bB)}_{\bb^j}\}$.  Now for the
matrix $\bB_{-1,\bi}=[\bi^T, (\bb^2)^T, \ldots , (\bb^b)^T]^T$ we have
\begin{eqnarray*}
\lefteqn{
\left|\det\left[\bB_{-1,\bi}\bB_{-1,\bi}^T\right]\right|} \nonumber \\*
&=&\left|
\det\left[\bB_{-1,\bi}\bi^T,\bB_{-1,\bi}(\bb^2)^T,\bB_{-1,\bi}(\bb^3)^T,\ldots, \bB_{-1,\bi}(\bb^b)^T\right]\right| \nonumber \\
&=&\left|\det\left[
\left(\sum_{j=1}^b \alpha_{\bb^j}^{(\bi,\bB)} \bB_{-1,\bi} \bb^j\right)^T,\bB_{-1,\bi}(\bb^2)^T,\bB_{-1,\bi}(\bb^3)^T,\ldots, 
\bB_{-1,\bi}(\bb^b)^T\right]\right| \nonumber \\
&=&\left|\sum_{j=1}^b \alpha_{\bb^j}^{(\bi,\bB)}
\det\left[\bB_{-1,\bi}(\bb^j)^T,\bB_{-1,\bi}(\bb^2)^T, \bB_{-1,\bi}(\bb^3)^T, \ldots,
  \bB_{-1,\bi}(\bb^b)^T\right]\right| \nonumber \\
&=&|\alpha_{\bb^1}^{(\bi,\bB)}|\left| \det\left[\bB_{-1,\bi}\bB^T\right]\right| \nonumber \\
&=&\left(\alpha_{\bb^1}^{(\bi,\bB)}\right)^2 \left|
\det\left[\bB\bB^T\right]\right |
\end{eqnarray*}
where last equality follows by the same argument the penultimate
equality was obtained. Repeating the same argument  for
$\bB_{-j,\bi}$, $j=2,\ldots,b$ we obtain
\begin{equation}
\left|\det\left[\bB_{-j,\bi}\bB_{-j,\bi}^T\right]\right| =
\left(\alpha_{\bb^j}^{(\bi,\bB)}\right)^2 \left|
\det\left[\bB\bB^T\right]\right |.
\label{eq:bary}
\end{equation}
Thus the maximal property of $|\det[\bB\bB^T]|$ implies
$|\alpha_{\bb^j}^{(\bi,\bB)}| \le 1$ for all $j=1,\ldots,b$.  The
second statement follows trivially from (\ref{eq:cspanner}) and
(\ref{eq:bary}).  \qed

Awerbuch and Kleinberg \cite{AwKl04} also present an iterative
algorithm to find a $C$-barycentric spanner if $\bB$ is a square
matrix. Starting from the identity matrix, their algorithm replaces a
row of the matrix in each step by maximizing the determinant with
respect to the given row. This is done by calling an oracle function,
and it is shown that the oracle is called $O(b \log_C b)$ times. In
case $\bB$ is not a square matrix, the algorithm carries over if we
have access to an alternative oracle that can maximize $|\det [\bB
\bB^T]|$: Starting from an arbitrary basis
$\bB$ we can iteratively replace one row in each step, using the
oracle, to maximize the determinant $|\det [\bB\bB^T]|$ until
(\ref{eq:cspanner}) is satisfied for all $j$ and $\bi$.  By
Lemma~\ref{le:bary}, this results in a $C$-barycentric spanner.
Similarly to \cite{AwKl04}, it can be shown that the oracle is called 
$O(b \log_C b)$ times for $C>1$.

For simplicity (to avoid carrying the constant $C$), assume that we
have a 2-barycentric spanner $\bB$.  Based on the ideas of label
efficient prediction, the next algorithm gives a simple solution to
the restricted shortest path problem.  The algorithm is very similar
to that of the algorithm in the label efficient case, but here we
cannot estimate the edge losses directly. Therefore, we query the loss
of a (random) basis vector from time to time, and create  unbiased
estimates $\ti{\ell}_{\bb^j,t}$ of the losses of basis paths
$\ell_{\bb^j,t}$, which are then transformed into edge-loss estimates.

\bookbox{
\begin{center}
A BANDIT ALGORITHM FOR THE RESTRICTED SHORTEST PATH PROBLEM
\end{center}
{\bf Parameters:} $0<\eps,\eta\le1$.\\
 {\bf Initialization:} Set $w_{\e,0}=1$ for
each $\e \in \E$, $\bw_{\bi,0}=1$ for each $\bi \in \cP$, $\bW_0=N$.
Fix a  basis $\bB$, which is a 2-barycentric spanner.
For each round $t=1,2,\ldots$
\begin{itemize}
\item[(a)] Draw a Bernoulli random variable $S_{t}$ such that $\PROB{(S_{t}=1)}=\eps$;
\item[(b)] If $S_{t}=1$, then choose the path $\bI_t$ uniformly
from the basis $\bB$. If $S_{t}=0$, then choose $\bI_t$
according to the distribution $\{p_{\bi,t}\}$, defined by
\[
p_{\bi,t}= \frac{\bw_{\bi,t-1}}{\bW_{t-1}}~.
\]
\item[(c)] Calculate the estimated loss of all edges according to
\[
\tilde{\bl}^{(\E)}_{t}=\bB^+\tilde{\bl}^{(\bB)}_{t}~,
\]
where $\tilde{\bl}^{(\E)}_{t}=\{\tilde{\ell}^{(\E)}_{e,t}\}_{e\in\E}$,
and
$\tilde{\bl}^{(\bB)}_{t}=(\tilde{\ell}^{(\bB)}_{\bb^1,t},\ldots,\tilde{\ell}^{(\bB)}_{\bb^b,t})$
is the vector of the estimated losses 
\[
\tilde{\ell}_{\bb^j, t} =
\frac{S_{t}}{\eps}\ell_{\bb^j,t}
\I_{\{\bI_t=\bb^j\}}b
\]
for $j=1,\ldots,b$.
\item[(d)]
Compute the updated weights
\begin{eqnarray*}
w_{\e,t}&=&w_{\e,t-1} e^{-\eta \tilde{\ell}_{e,t}}, \\
\bw_{\bi,t}&=& \prod_{\e \in \bi} w_{\e,t} =\bw_{\bi,t-1} e^{-\eta
\sum_{\e \in \bi} \tilde{\ell}_{e,t}},
\end{eqnarray*}
and the sum of the total weights of the paths
\[
\bW_{t}=\sum_{\bi \in \cP} \bw_{\bi,t}~.
\]
\end{itemize}
}

The performance of the algorithm is analyzed in the next theorem. The
proof follows the argument of Cesa-Bianchi \emph{et al$.$}
\cite{CeLuSt05}, but we also have to deal with some additional
technical difficulties. Note that in the theorem we do not assume that
all paths between $u$ and $v$ have equal length.

\begin{theorem}
\label{th:lepathbandit}
Let $K$ denote the length of the longest path in the graph.  For any
$\delta \in (0,1)$, parameters $0 < \eps \le \frac{1}{K}$ and $\eta>0$
satisfying $\eta \le \eps^2$, and $n \ge \frac{8b}{\eps^2} \ln
\frac{4bN}{\d}$, the performance of the algorithm defined above can be
bounded, with probability at least $1-\delta$, as
\[
\wh{L}_n-\min_{\bi \in \cP} L_{\bi,n} 
\le 
K\left( \frac{\eta b}{\eps}Kn + \sqrt{\frac{n}{2} \ln \frac{4}{\d}} + n\eps
+\frac{\sqrt{ 2n\eps\ln \frac{4}{\d}}}{K}
+\frac{16}{3}b\sqrt{2n\frac{b}{\eps}\ln \frac{4bN}{\d}}\right) + \frac{\ln N}{\eta}
\]
In particular, choosing
\[
\eps= \left(\frac{Kb}{n}\ln \frac{4bN}{\d}\right)^{1/3}
\quad \text{ and } \quad \eta = \eps^2 
\]
we obtain
\[
\wh{L}_n - \min_{\bi \in \cP} L_{\bi,n} \le
9.1 K^2b\left(Kb\ln (4bN/\d)\right)^{1/3}n^{2/3}~.
\]
\end{theorem}

\bigskip

The theorem is proved using the following two lemmas.
The first one is an easy consequence of Bernstein's inequality:
\begin{lemma}\label{le:binomial}
Under the assumptions of Theorem~\ref{th:lepathbandit}, the probability that the
algorithm queries the basis more than $n\eps+\sqrt{ 2n\eps
\ln \frac{4}{\d}}$ times is at most $\delta/4$.
\end{lemma}

Using the estimated loss of a path $\bi\in\cP$ given in
(\ref{eq:pathestimate}), we can estimate the cumulative loss of $\bi$ up to
time $n$ as
\[
\tilde{L}_{\bi,n}=\sum_{t=1}^n \tl_{\bi,t}~.
\]
The next lemma demonstrates the quality of these estimates.

\begin{lemma}
\label{le:LE-berstein}
Let $ 0 < \d <1$ and assume
 $n\ge \frac{8b}{\eps} \ln \frac{4bN}{\d}$.
For any $\bi \in \cP$, with probability at least $1-\d/4$,
\[
\sum_{t=1}^{n}\sum_{\bi \in \cP} p_{\bi,t} \ell_{\bi,t}-
\sum_{t=1}^{n}\sum_{\bi \in \cP} p_{\bi,t} \tl_{\bi,t}
\le \frac{8}{3}b\sqrt{2n\frac{bK^2}{\eps}\ln \frac{4b}{\d}}~.
\]
Furthermore, with probability at least $1-\d/(4N)$,
\[
\ti{L}_{\bi,n} -  L_{\bi,n}
\le \frac{8}{3}b\sqrt{2n\frac{bK^2}{\eps}\ln \frac{4bN}{\d}}~.
\]
\end{lemma}
\proof
We may write
\begin{eqnarray}
\sum_{t=1}^{n}\sum_{\bi \in \cP} p_{\bi,t} \ell_{\bi,t}-
\sum_{t=1}^{n}\sum_{\bi \in \cP} p_{\bi,t} \tl_{\bi,t}
&=&\sum_{t=1}^{n} \sum_{\bi \in \cP} p_{\bi,t}
\sum_{j=1}^b\alpha_{\bb^j}^{(\bi,\bB)} 
\left(\ell_{\bb^j,t}-\tl_{\bb^j,t}\right) \notag \\*
&=&\sum_{j=1}^b \sum_{t=1}^{n} \left[\sum_{\bi \in \cP} p_{\bi,t}
  \alpha_{\bb^j}^{(\bi,\bB)} 
\left(\ell_{\bb^j,t}-\tl_{\bb^j,t}\right)\right] \notag\\*
&\defeq&\sum_{j=1}^b\sum_{t=1}^{n} X_{\bb^j,t}~. \label{eq:bernstein1}
\end{eqnarray}
Note that for any $\bb^j$, $X_{\bb^j,t}$, $t=1,2,\ldots$ is a
martingale difference sequence with respect to $(\bI_t,S_t)$,
$t=1,2,\ldots$  as $\EXP_t \ti{\ell}_{\bb,t}=\ell_{\bb,t}$. Also,
\begin{eqnarray}
\EXP_{t}[X^2_{\bb^j,t}] 
\le  \left(\sum_{\bi \in \cP} p_{\bi,t} \alpha_{\bb^j}^{(\bi,\bB)}\right)^2
\EXP_{t}\left[\left(\tl_{\bb^j,t} \right)^2\right] 
\le  \sum_{\bi \in \cP} p_{\bi,t} \left( \alpha_{\bb^j}^{(\bi,\bB)}\right)^2
\frac{K^2b}{\eps} 
\le  4 \frac{K^2b}{\eps} \label{eq:bern_var}
\end{eqnarray}
and
\begin{equation}
\label{eq:bern_abs2}
|X_{\bb^j,t}| \le \left|\sum_{\bi \in \cP} p_{\bi,t} \alpha_{\bb^j}^{(\bi,\bB)}\right| \left|\ell_{\bb^j,t} -\tl_{\bb^j,t} \right|
\le \sum_{\bi \in \cP} p_{\bi,t} \left|\alpha_{\bb^j}^{(\bi,\bB)}\right| \frac{Kb}{\eps}
\le 2\frac{Kb}{\eps} 
\end{equation}
where the last inequalities in both cases follow from the fact that $\bB$
is a $2$-barycentric spanner. 
Then, using Bernstein's inequality for martingale
differences (Lemma \ref{le:Bernstein}), we have, for any fixed $\bb^j$,
\begin{equation}
\PROB\left[\sum_{t=1}^n X_{\bb^j,t} \ge
\frac{8}{3}\sqrt{2n\frac{bK^2}{\eps}\ln \frac{4b}{\d}}\right] \le \frac{\d}{4b} \label{eq:bernstein2}
\end{equation} 
where we used (\ref{eq:bern_var}), (\ref{eq:bern_abs2}) and the assumption of
the lemma on $n$. The proof of the first statement is finished with an application
of the union bound and its combination with (\ref{eq:bernstein1}).

For the second statement we use a similar argument, that is,
\begin{eqnarray}
\sum_{t=1}^n (\tl_{\bi,t}-\ell_{\bi,t})
=\sum_{j=1}^b \alpha_{\bb^j}^{(\bi,\bB)}  \sum_{t=1}^n
(\tl_{\bb^j,t}-\ell_{\bb^j,t}) 
&\le& \sum_{j=1} \left|\alpha_{\bb^j}^{(\bi,\bB)}\right| \left|\sum_{t=1}^n  (\tl_{\bb^j,t}-\ell_{\bb^j,t})\right| \notag\\*
&\le& 2\sum_{j=1}^b \left|\sum_{t=1}^n  (\tl_{\bb^j,t}-\ell_{\bb^j,t})\right|~. \label{eq:bernstein3}
\end{eqnarray}
Now applying Lemma \ref{le:Bernstein} for a fixed $\bb^j$ we get
\begin{equation}
\PROB\left[\sum_{t=1}^n (\tl_{\bb^j,t}-\ell_{\bb^j,t}) \ge
\frac{4}{3} \sqrt{2n\frac{K^2b}{\eps}\ln \frac{4bN}{\d}}
\right]\le \frac{\d}{4bN} \label{eq:bernstein4}
\end{equation}
because of $\EXP_{t}[(\tl_{\bb^j,t}-\ell_{\bb^j,t})^2] \le \frac{K^2b}{\eps}$
and $ -K \le \tl_{\bb^j,t}-\ell_{\bb^j,t} \le K\left(\frac{b}{\eps}-1\right)$.
The proof is completed by applying the  union bound   to
(\ref{eq:bernstein4}) and combining the result with (\ref{eq:bernstein3}).
\qed

\proofof{Theorem~\ref{th:lepathbandit}} Similarly to earlier
proofs, we follow the evolution of the term $\ln
\frac{\bW_{n}}{\bW_{0}}$.  In the same way as we obtained
(\ref{eq:also}) and (\ref{eq:felso}), we have
\begin{eqnarray*}
\ln \frac{\bW_{n}}{\bW_{0}} \ge -\eta \min_{\bi \in \cP} \tL_{\bi,n} - \ln N
\end{eqnarray*}
and
\begin{eqnarray*}
\ln \frac{\bW_{n}}{\bW_{0}} &\le&  \sum_{t=1}^n \left( -\eta \sum_{\bi\in\cP}p_{\bi,t}\tl_{\bi,t}
+\frac{\eta^2}{2} \sum_{\bi\in\cP}p_{\bi,t}\tl_{\bi,t}^2\right)~.
\end{eqnarray*}
Combining these bounds, we obtain
\begin{eqnarray*}
-\min_{\bi \in \cP} \tL_{\bi,n} - \frac{\ln N}{\eta} &\le& \sum_{t=1}^n \left( -\sum_{\bi\in\cP}p_{\bi,t}\tl_{\bi,t}
+\frac{\eta}{2} \sum_{\bi\in\cP}p_{\bi,t}\tl_{\bi,t}^2\right)\\*
&\le& \left(-1+\frac{\eta Kb}{\eps}\right)\sum_{t=1}^n  \sum_{\bi\in\cP}p_{\bi,t}\tl_{\bi,t}~,
\end{eqnarray*}
because $0 \le \tl_{\bi,t}\le \frac{2Kb}{\eps}$.
Applying the results of Lemma \ref{le:LE-berstein} and the union bound, we have, with
probability $1-\d/2$,
\begin{eqnarray}
\lefteqn{-\min_{\bi \in \cP} L_{\bi,n}-\frac{8}{3}b\sqrt{2n\frac{K^2b}{\eps}\ln \frac{4bN}{\d}}} \notag\\*
&\le& \left(-1+\frac{\eta Kb}{\eps}\right)\left(\sum_{t=1}^n\sum_{\bi\in\cP}p_{\bi,t}\ell_{\bi,t}
-\frac{8}{3}b\sqrt{2n\frac{K^2b}{\eps}\ln \frac{4b}{\d}}\right)+ \frac{\ln N}{\eta} \notag\\*
&\le& \left(-1+\frac{\eta Kb}{\eps}\right) \sum_{t=1}^n\sum_{\bi\in\cP}p_{\bi,t}\ell_{\bi,t}
+ \frac{8}{3}b\sqrt{2n\frac{K^2b}{\eps}\ln \frac{4b}{\d}}+ \frac{\ln N}{\eta}~. \label{eq:1-d/2}
\end{eqnarray}
Introduce the sets
\[
    \mc{T}_n \defeq \{t: 1\le t \le n \text{ and } S_t=0\} \qquad \text{and} \qquad
    \ol{\mc{T}}_n \defeq \{t: 1\le t \le n \text{ and } S_t=1\}~
\]
of ``exploitation'' and ``exploration'' steps, respectively. Then, by
the Hoeffding-Azuma inequality \cite{Hoe63} we obtain that,
with probability at least $1-\d/4$,
\[
\sum_{t \in \mc{T}_n} \sum_{\bi \in \cP} p_{\bi,t} \ell_{\bi,t} \ge \sum_{t\in\mc{T}_n} \ell_{\bI_t,t}
- \sqrt{\frac{|\mc{T}_n|K^2}{2} \ln \frac{4}{\d}}~.
\]
Note that for the exploration steps $t\in\ol{\mc{T}}_n$, as the algorithm
plays according to a uniform distribution instead of $p_{\bi,t}$,
we can only use the trivial lower bound zero on the losses,
that is,
\[
\sum_{t \in \ol{\mc{T}}_n} \sum_{\bi \in \cP} p_{\bi,t} \ell_{\bi,t} \ge \sum_{t\in\ol{\mc{T}}_n} \ell_{\bI_t,t} - K|\ol{\mc{T}}_n|~.
\]
The last two inequalities imply
\begin{equation}
\sum_{t =1}^n \sum_{\bi \in \cP} p_{\bi,t} \ell_{\bi,t} \ge \wh{L}_n
- \sqrt{\frac{|\mc{T}_n|K^2}{2} \ln \frac{4}{\d}}- K|\ol{\mc{T}}_n|~. \label{eq:hoeff}
\end{equation}
Then, by (\ref{eq:1-d/2}), (\ref{eq:hoeff}) and Lemma \ref{le:binomial} we obtain, with probability at least $1-\d$,
\begin{eqnarray*}
\lefteqn{\wh{L}_n-\min_{\bi \in \cP} L_{\bi,n}}\\*
&\le& K\left( \frac{\eta b}{\eps}Kn + \sqrt{\frac{n}{2} \ln \frac{4}{\d}} + n\eps
+ \frac{\sqrt{ 2n\eps\ln \frac{4}{\d}}}{K}
+\frac{16}{3}b\sqrt{2n\frac{b}{\eps}\ln \frac{4bN}{\d}}\right) + \frac{\ln N}{\eta}
\end{eqnarray*}
where we used $\wh{L}_n\le Kn$ and $|\mc{T}_n|\le n$.
Substituting the values of $\eps$ and $\eta$ gives
\begin{eqnarray*}
\wh{L}_n-\min_{\bi \in \cP} L_{\bi,n}
&\le& K^2bn\eps + \frac{1}{4}Kn\eps + Kn\eps
+\frac{1}{2}n\eps
+\frac{16}{3}b \sqrt{K}n\eps + n\eps\\*
&\le& 9.1 K^2bn\eps
\end{eqnarray*}
where we used
$\sqrt{ \frac{n}{2} \ln \frac{4}{\d}} \le \frac{1}{4}n\eps$,
$\sqrt{ 2n\eps\ln \frac{4}{\d}} \le \frac{1}{2}n\eps$,
$\sqrt{n\frac{bK}{\eps}\ln \frac{4N}{\d}}=n\eps$, and
 $\frac{\ln N}{\eta} \le n\eps$
(from  the assumptions of the theorem).
\qed

\section{Simulation results}
\label{sec:sim} 

To further investigate our new algorithms, we have conducted some simple
simulations. As the main motivation of this work is to improve earlier
algorithms in case the number of paths is exponentially large in the
number of edges, we tested the algorithms on the small graph shown in
Figure~\ref{f:dags} (b), which has one of the simplest structures with
exponentially many (namely $2^{|\E|/2}$) paths.

\begin{figure}
\begin{center}
\includegraphics[angle=0, width=\minipagewidth]{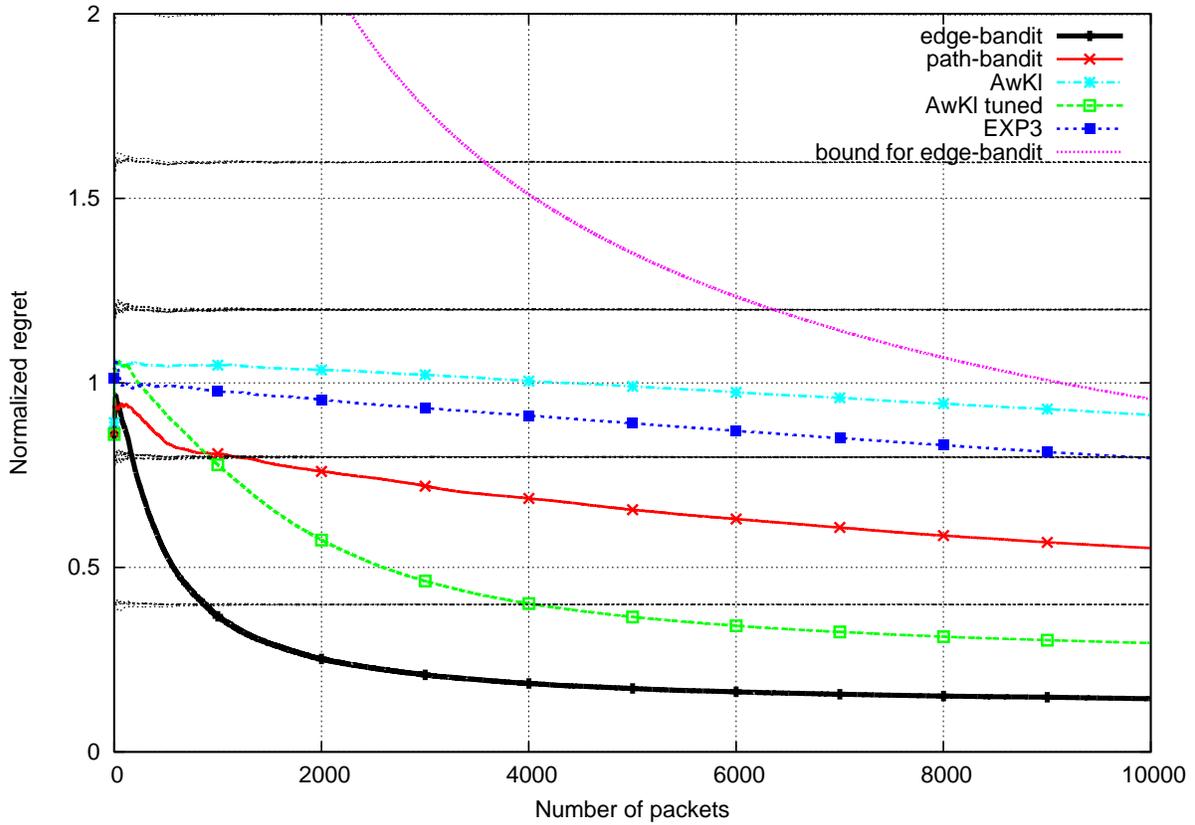}
\end{center}
\caption{Normalized regret of several algorithms for the shortest path
problem. The gray dotted lines show the normalized regret of fixed paths in the graph. }
\label{simulation}
\end{figure}

The losses on the edges were generated by a sequence of  independent
and uniform random variables, with values from $[0,1]$ on the upper
edges, and from $[0.32,1]$ on the lower edges, resulting in a
(long-term) optimal 
path consisting of the upper edges. We ran the tests for $n=10000$
steps, with confidence value $\delta=0.001$.  To establish baseline
performance, we also tested the EXP3 algorithm of Auer \emph{et al$.$}
\cite{AuCeFrSc02} (note that this algorithm does not need edge losses,
only the loss of the chosen path).  For the version of our bandit
algorithm that is informed of the individual edge losses
(edge-bandit), we used the simple 2-element cover set of the paths
consisting of the upper and lower edges, respectively (other 2-element
cover sets give similar performance). For our restricted shortest path
algorithm (path-bandit) the basis $\{uuuuu, uuuul, uuull,uulll,
ullll\}$ was used, where $u$ (resp.\ $l$) in the $k$th position
denotes the upper (resp.\ lower) edge connecting $v_{k-1}$ and
$v_k$. In this example the performance of the algorithm appeared to be
independent of the actual choice of the basis; however, in general we
do not expect this behavior.  Two versions of the algorithm of
Awerbuch and Kleinberg \cite{AwKl04} were also simulated.  With its
original parameter setting (AwKl), the algorithm did not perform
well. However, after optimizing its parameters off-line (AwKl tuned),
substantially better performance was  achieved.  The normalized
regret of the above algorithms, averaged over 30 runs, as well as the
regret of the fixed paths in the graph are shown in
Figure~\ref{simulation}.

Although all algorithms showed better performance than the bound for
the edge-bandit algorithm, the latter showed the expected superior
performance  in the simulations. Furthermore, our algorithm for
the restricted shortest path problem outperformed Awerbuch and
Kleinberg's (AwKl) algorithm, while being inferior to its off-line
tuned version (AwKl tuned). It must be noted that similar parameter
optimization did not improve the performance of our path-bandit
algorithm, which showed robust behavior with respect to parameter
tuning.

\section{Conclusions}
\label{sec:conc} \vspace{-0.2cm}

We considered different versions of the on-line shortest path problem 
with limited feedback.
These problems are motivated by realistic scenarios, such as  routing
in  communication networks, where the vertices do not have all the
information about the state of the network. We have addressed the
problem in the adversarial setting where the edge losses may vary
in an arbitrary way; in particular, they may depend on previous
routing decisions of the algorithm. Although this assumption may
neglect natural correlation in the loss sequence, it suits
applications in mobile ad-hoc networks, where the network topology
changes dynamically in time, and also in certain secure networks
that has to be able to handle denial of service attacks.

Efficient algorithms have been provided for 
the multi-armed bandit setting and in a combined
label efficient multi-armed bandit setting, provided the individual
edge losses along the chosen path are revealed to the algorithms. 
The normalized regrets of the
algorithms, compared to the performance of the best fixed path,
converge to zero at an $O(1/\sqrt{n})$ rate as the time horizon $n$
grows to infinity, and increases only polynomially in the number of
edges (and vertices) of the graph. Earlier methods for the multi-armed
bandit problem either do not have the right $O(1/\sqrt{n})$
convergence rate, or their regret increase exponentially in the
number of edges for typical graphs. The algorithm has also been 
extended so that it can  compete with time varying paths, that is, 
to handle situations when the best path can change from time to time 
(for consistency, the number of changes must be sublinear in $n$). 

In the restricted version of the shortest path problem, where only the
losses of the whole paths are revealed, an algorithm with a 
worse $O(n^{-1/3})$ normalized regret was provided.  This algorithm
has comparable performance to that of the best earlier algorithm for
this problem \cite{AwKl04}, however, our algorithm is significantly
simpler.  Simulation results are also given to assess the practical
performance and compare it to the theoretical bounds as well as other
competing algorithms.

It should be noted that the results are not entirely satisfactory in
the restricted version of the problem, as it remains  an open
question whether the $O(1/\sqrt{n})$ regret can be achieved without
the exponential dependence on the size of the graph.  Although we
expect that this is the case, we have not been able to construct an
algorithm with such a proven  performance bound.

\section{Appendix}
\begin{lemma} \label{le:Bernstein}
(Bernstein's inequality for martingale differences \cite{Fre75}.)
Let $X_1,\ldots,X_n$ be a martingale difference sequence such that
$X_t \in [a,b]$ with probability one ($t=1, \ldots,n$). Assume that,
for all $t$,
\[
    \EXP \left[X_t^2|X_{t-1},\ldots,X_1 \right]\leq \sigma^2 \text{ }a.s.
\]
Then, for all $\eps > 0$,
\[
    \mathbb{P} \left\{ \sum_{t=1}^n X_t  > \eps\right\} \le e^{\frac{-\eps^2}{2n\sigma^2+2\eps(b-a)/3} }
\]
and therefore
\[
   \mathbb{P} \left\{ \sum_{t=1}^n X_t  > \sqrt{2n\sigma^2 \ln \delta^{-1}} + 2\ln \delta^{-1}(b-a)/3 \right\} \le \delta.
\]
\end{lemma}


\begin{thebibliography}{10}

\bibitem{AuCeFrSc02}
P.~Auer, N.~Cesa-Bianchi, Y.~Freund, and R.~Schapire.
\newblock The non-stochastic multi-armed bandit problem.
\newblock {\em SIAM Journal on Computing}, 32(1):48--77, 2002.

\bibitem{AW98}
P.~Auer and M.~K. Warmuth, ``Tracking the best disjunction,'' {\em Machine
  Learning}, vol.~32, no.~2, pp.~127--150, 1998.

\bibitem{AwHoRuKl05}
B.~Awerbuch, D.~Holmer, H.~Rubens, and R.~Kleinberg.
\newblock Provably competitive adaptive routing.
\newblock In {\em Proceedings of IEEE INFOCOM 2005}, volume~1, pages 631--641,
  March 2005.

\bibitem{AwKl04}
B.~Awerbuch and R.~D. Kleinberg.
\newblock Adaptive routing with end-to-end feedback: distributed learning and
  geometric approaches.
\newblock In {\em Proceedings of the 36th Annual ACM Symposium on the Theory of
  Computing, STOC 2004}, pages 45--53, Chicago, IL, USA, Jun. 2004. ACM Press.

\bibitem{Bla56}
D.~Blackwell.
\newblock An analog of the minimax theorem for vector payoffs.
\newblock {\em Pacific Journal of Mathematics}, 6:1--8, 1956.

\bibitem{BoWa02}
O.~Bousquet and M.~K. Warmuth.
\newblock Tracking a small set of experts by mixing past posteriors.
\newblock {\em Journal of Machine Learning Research}, 3:363--396, Nov. 2002.

\bibitem{CeFrHaHeScWa97}
N.~Cesa-Bianchi, Y.~Freund, D.~P. Helmbold, D.~Haussler, R.~Schapire, and M.~K.
  Warmuth.
\newblock How to use expert advice.
\newblock {\em Journal of the ACM}, 44(3):427--485, 1997.

\bibitem{CeLu06}
N.~Cesa-Bianchi and G.~Lugosi.
\newblock {\em Prediction, Learning, and Games}.
\newblock Cambridge University Press, Cambridge, 2006.

\bibitem{CeLuSt05}
N.~Cesa-Bianchi, G.~Lugosi, and G.~Stoltz.
\newblock Minimizing regret with label efficient prediction.
\newblock {\em IEEE Trans. Inform. Theory}, IT-51:2152--2162, June 2005.


\bibitem{Fre75} 
D.~A.~Freedman.
\newblock On tail probabilities for martingales.
\newblock {\em Annals of Probability}, 3:100--118, Feb.\ 1975.

\bibitem{GeGeLeLiSu04}
E.~Gelenbe, M.~Gellman, R.~Lent, P.~Liu, and P.~Su.
\newblock Autonomous smart routing for network {QoS}.
\newblock In {\em Proceedings of First International Conference on Autonomic
  Computing}, pages 232--239, New York, May 2004. IEEE Computer Society.

\bibitem{GeLeXh01}
E.~Gelenbe, R.~Lent, and Z.~Xhu.
\newblock Measurement and performance of a cognitive packet network.
\newblock {\em Journal of Computer Networks}, 37:691--701, 2001.

\bibitem{GyLiLu04}
A.~Gy\"orgy, T.~Linder, and G.~Lugosi.
\newblock Efficient algorithms and minimax bounds for zero-delay lossy source
  coding.
\newblock {\em IEEE Transactions on Signal Processing}, 52:2337--2347, Aug.
  2004.

\bibitem{GyLiLu04a}
A.~Gy\"orgy, T.~Linder, and G.~Lugosi.
\newblock A "follow the perturbed leader"-type algorithm for zero-delay
  quantization of individual sequences.
\newblock In {\em Proc. Data Compression Conference}, pages 342--351, Snowbird,
  UT, USA, Mar. 2004.

\bibitem{GyLiLu05a}
A.~Gy\"orgy, T.~Linder, and G.~Lugosi.
\newblock Tracking the best of many experts.
\newblock In {\em Proceedings of the 18th Annual Conference on Learning Theory,
  COLT 2005}, pages 204--216, Bertinoro, Italy, Jun. 2005. Springer.

\bibitem{GyLiLu05d}
A.~Gy\"orgy, T.~Linder, and G.~Lugosi.
\newblock Tracking the best quantizer.
\newblock In {\em Proceedings of the IEEE International Symposium on
  Information Theory}, pages 1163--1167, Adelaide, Australia, June-July 2005.

\bibitem{GyOt06}
A.~Gy{\"o}rgy and {Gy}. Ottucs{\'a}k.
\newblock Adaptive routing using expert advice.
\newblock {\em The Computer Journal}, 49(2):180--189, 2006.

\bibitem{Han57}
J.~Hannan.
\newblock Approximation to Bayes risk in repeated plays.
\newblock In M.~Dresher, A.~Tucker, and P.~Wolfe, editors, {\em Contributions
  to the Theory of Games}, volume~3, pages 97--139. Princeton University Press,
  1957.

\bibitem{HeWa98}
M.~Herbster and M.~K. Warmuth.
\newblock Tracking the best expert.
\newblock {\em Machine Learning}, 32(2):151--178, 1998.

\bibitem{HeWa01}
M.~Herbster and M.~K. Warmuth, ``Tracking the best linear predictor,'' {\em
  Journal of Machine Learning Research}, vol.~1, pp.~281--309, 2001.

\bibitem{Hoe63}
W.~Hoeffding.
\newblock Probability inequalities for sums of bounded random variables.
\newblock {\em Journal of the American Statistical Association}, 58:13--30,
  1963.

\bibitem{HP97}
D.P. Helmbold and S.~Panizza.
\newblock Some label efficient learning results.
\newblock In {\em Proceedings of the 10th Annual Conference on Computational
  Learning Theory}, pages 218--230. ACM Press, 1997.

\bibitem{KaVe03}
A.~Kalai and S~Vempala.
\newblock Efficient algorithms for the online decision problem.
\newblock In B.~Sch\"olkopf and M.~Warmuth, editors, {\em Proceedings of the
  16th Annual Conference on Learning Theory and the 7th Kernel Workshop,
  COLT-Kernel 2003}, pages 26--40, New York, USA, Aug. 2003. Springer.

\bibitem{LiWa94}
N.~Littlestone and M.~K. Warmuth.
\newblock The weighted majority algorithm.
\newblock {\em Information and Computation}, 108:212--261, 1994.

\bibitem{McBl04}
H.~B. McMahan and A.~Blum.
\newblock Online geometric optimization in the bandit setting against an
  adaptive adversary.
\newblock In {\em Proceedings of the 17th Annual Conference on Learning Theory,
  COLT 2004}, pages 109--123, Banff, Canada, Jul. 2004. Springer.

\bibitem{moh98}
M.~Mohri.
\newblock General algebraic frameworks and algorithms for shortest distance
  problems.
\newblock Technical Report 981219-10TM, AT{\&}T Labs Research, 1998.

\bibitem{HeSc97}
R.~E. Schapire and D.~P. Helmbold.
\newblock Predicting nearly as well as the best pruning of a decision tree.
\newblock {\em Machine Learning}, 27:51--68, 1997.

\bibitem{TaWa03}
E.~Takimoto and M.~K. Warmuth.
\newblock Path kernels and multiplicative updates.
\newblock {\em Journal of Machine Learning Research}, 4:773--818, 2003.

\bibitem{Vov90}
V.~Vovk.
\newblock Aggregating strategies.
\newblock In {\em Proceedings of the Third Annual Workshop on Computational
  Learning Theory}, pages 372--383, Rochester, NY, Aug. 1990. Morgan Kaufmann.

\bibitem{Vov99}
V.~Vovk.
\newblock Derandomizing stochastic prediction strategies.
\newblock {\em Machine Learning}, 35(3):247--282, Jun. 1999.

\end{thebibliography}
\end{document}